\documentclass[runningheads]{llncs}
\usepackage{graphicx}
\usepackage{amsmath}
\usepackage{amssymb}
\usepackage{array}
\usepackage{subcaption}
\usepackage{hyperref}
\usepackage{xcolor}
\usepackage{ulem}
\usepackage{multirow}
\usepackage{booktabs}
\usepackage{bm}
\usepackage{tabularx}
\usepackage{float}
\usepackage{placeins}
\usepackage{svg}

\newcolumntype{C}[1]{>{\centering\arraybackslash}p{#1}}

\begin{document}

\title{Arbitrary Data as Images: Fusion of Patient Data Across Modalities and Irregular Intervals with Vision Transformers}
\titlerunning{Fusion of Patient Data Across Modalities and Irregular Intervals}

\author{
    Malte T\"olle\inst{*,1,2,3} \and
    Mohamad Scharaf\inst{*,1} \and
    Samantha Fischer\inst{1,2,3} \and 
    Christoph Reich\inst{1,2,3} \and
    Silav Zeid\inst{4,5,6} \and
    Christoph Dieterich\inst{1,2,3} \and
    Benjamin Meder\inst{1,2,3} \and
    Norbert Frey\inst{1,2,3} \and 
    Philipp Wild\inst{4,5,6,7} \and 
    Sandy~Engelhardt\inst{1,2,3}
}
\authorrunning{M. T\"olle et al.}
%
\institute{
    Department of Cardiology, Angiology and Pneumology, Heidelberg University Hospital, Heidelberg, Germany \and
    Informatics for Life Institute, Heidelberg, Germany \and
    DZHK (German Centre for Cardiovascular Research), partner site Heidelberg/Mannheim, Heidelberg, Germany \and
    Preventive Cardiology and Preventive Medicine, Department of Cardiology, University Medical Center of the Johannes Gutenberg University Mainz, Mainz, Germany \and
    Clinical Epidemiology and Systems Medicine, Center for Thrombosis and Hemostasis, University Medical Center Mainz, Johannes Gutenberg University Mainz, Germany \and
    DZHK (German Centre for Cardiovascular Research), partner site Rhine-Main, Mainz, Germany \and
    Systems Medicine, Institute of Molecular Biology (IMB), Mainz, Germany \\
    * contributed equally \\
    \email{malte.toelle@med.uni-heidelberg.de}
}
\maketitle              
\begin{abstract}
    A patient undergoes multiple examinations in each hospital stay, where each provides different facets of the health status.
    These assessments include  temporal data with varying sampling rates, discrete single-point measurements, therapeutic interventions such as medication administration, and images. 
    While physicians are able to process and integrate diverse modalities intuitively, neural networks need specific modeling for each modality complicating the training procedure.
    We demonstrate that this complexity can be significantly reduced by visualizing all information as images along with unstructured text and subsequently training a conventional vision-text transformer.
    Our approach, \textbf{Vi}sion \textbf{T}ransformer for \textbf{i}rregular sampled \textbf{M}ulti-modal \textbf{M}easurements (\textbf{ViTiMM}), not only simplifies data preprocessing and modeling but also outperforms current state-of-the-art methods in predicting in-hospital mortality and phenotyping, as evaluated on 6,175 patients from the MIMIC-IV dataset. 
    The modalities include patient's clinical measurements, medications, X-ray images, and electrocardiography scans. 
    We hope our work inspires advancements in multi-modal medical AI by reducing the training complexity to (visual) prompt engineering, thus lowering entry barriers and enabling no-code solutions for training.
    The source code will be made publicly available.
    
\end{abstract}

\setcounter{footnote}{0}

\section*{Introduction}

\begin{figure}[t]
    \centering
    \includegraphics[width=0.8\linewidth]{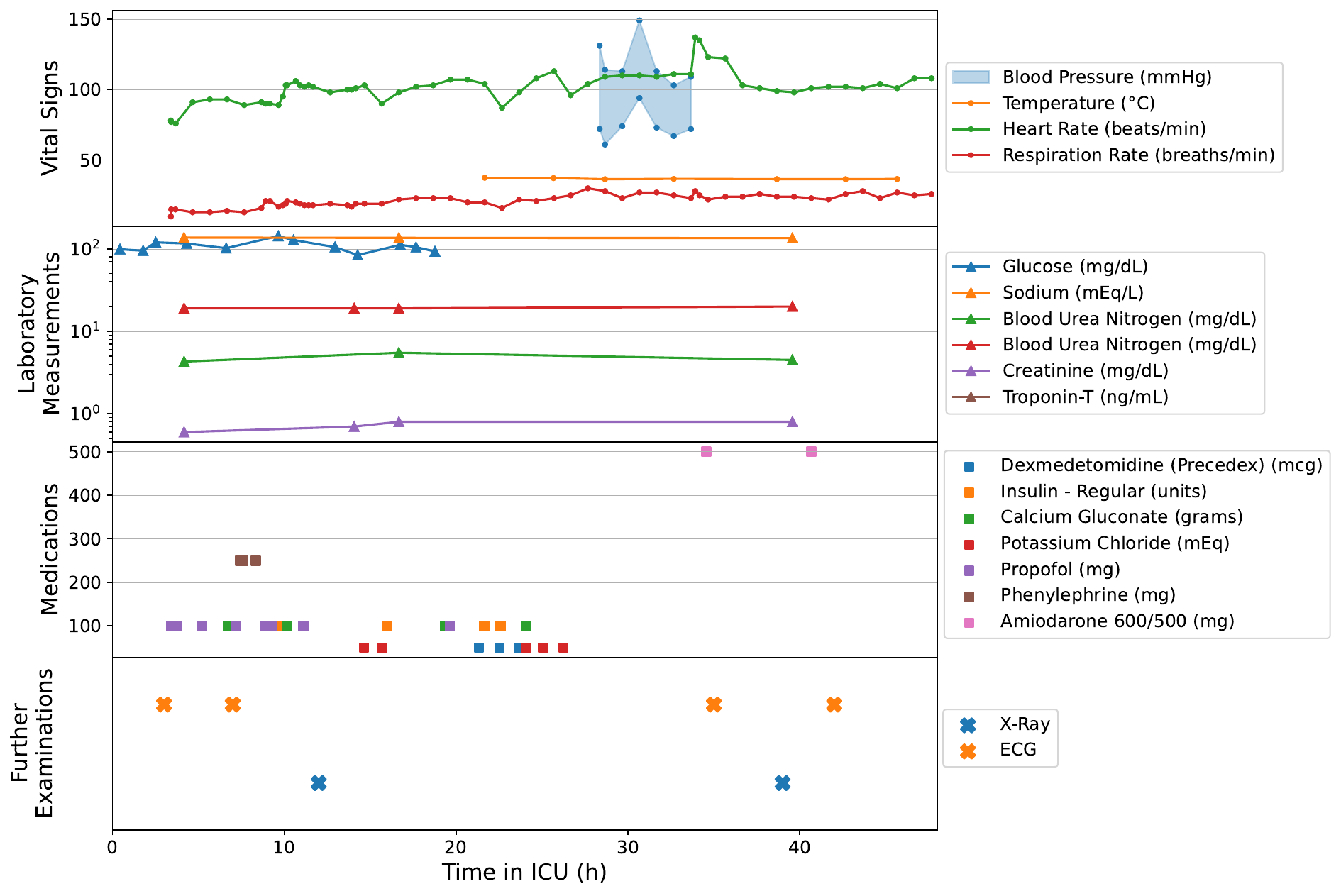}
    \caption{
        Example patient journey in the first 48 hours in the intensive care unit for \texttt{hadm\_id\,28016813} from the MIMIC-IV dataset~\cite{johnson2023mimiciv}.
        The patient examinations include, among others, vital signs (e.g., blood pressure), laboratory measurements (e.g., glucose), medications (e.g., insulin), and additional tests such as X-ray images and electrocardiography scans.
        The measurements are recorded at irregular time intervals, which differ between patients, adding further complexity to the modeling process.
        Additionally, medications can be administered in various doses, which must also be considered.
    }
    \label{fig:patient-journey}
\end{figure}

During a hospital stay, a patient typically undergoes multiple examinations, each offering distinct insights into their health status.
While physicians have learned to intuitively extract the different information and assemble them to an overall picture, neural networks need specific modeling of the different modalities and their interactions.
Nevertheless, once these challenges are addressed, multi-modal models have demonstrated promising performance~\cite{acosta2022multimodal,chambon2022roentgen,duan2021multimodal,huang2020multimodalb,huang2020multimodala,jabbour2022multimodal,Moor2023,touvron2023llama}.
However, a significant challenge persists: How to integrate multi-modal data that is captured at irregularly sampled time intervals (Fig.~\ref{fig:patient-journey})?


Patients admitted to the intensive care unit (ICU) require close monitoring and extensive diagnostics to restore a positive health status.
The extensive assessment of a patient's health leads to significant resources being spent on ICU patients, in the US alone they amount for up to 1\% of GPD annually~\cite{halpern2010icugdpus}.
Laboratory measurements are conducted at irregular time intervals to monitor vital organ functions, detect abnormalities, and guide treatment decisions. 
These tests provide critical information on parameters such as electrolyte balance, blood gases, and metabolic status. 
When a patient's values deviate from the ideal range, a variety of medications, often prepared in different solutions, are administered based on the patient's specific needs.
Other often but irregularly performed examinations include X-ray to monitor conditions such as heart failure or bone status and electrocardiography (ECG) scans to assess the electrical activity of the heart.
Physicians are then tasked with integrating all the partial information from different, irregular time points into a comprehensive picture to evaluate whether the patient requires additional care.

Multi-modal deep learning in medicine holds the potential to reduce the workload of physicians in such cases but faces a significant challenge: the integration of multiple modalities into a single model due to differing modeling requirements~\cite{acosta2022multimodal,kline2022multimodalsummary}.
Our primary contribution is a substantial reduction of the modeling complexity for multiple irregularly sampled modalities by transforming each modality into an image representation.
Humans are then tasked with visualizing the different modalities in an informative manner, effectively engaging in a form of "visual prompt engineering".
For example, laboratory measurements can be represented as line graphs over time to convey trends and patterns~\cite{li2023vitst}.
Our approach, \textbf{Vi}sion \textbf{T}ransformer for \textbf{i}rregular sampled \textbf{M}ulti-modal \textbf{M}easurements (\textbf{ViTiMM}), unifies the data processing pipeline, significantly reducing modeling complexity. 
This approach not only mimics the way humans interpret diverse data streams but also demonstrates significant improvements across a range of tasks.

In conventional approaches, each modality requires a distinct model architecture or embedding, which must then be fused to generate outputs~\cite{huang2020fusion}. 
For imaging modalities, features can be extracted using techniques such as convolutional neural networks and subsequently concatenated with features from another imaging modality~\cite{nejati2015imageimage,tawfik2022imageimage} or with tabular patient metadata~\cite{hyun2019imagemetadata,nie2019imagemetadata}.
When incorporating additional modalities, such as ECG, a separate model architecture must be utilized~\cite{mckeen2024ecgfm,ribeiro2020ecg}. 
In recent years, transformer models~\cite{vaswani2017attention} have simplified the modeling process, as the attention mechanism provides a general framework for handling diverse modalities~\cite{dosovitskiy2021vit,mckeen2024ecgfm}.
Despite this, different modalities still require unique embedding mechanisms, and careful use of cross-attention blocks is necessary to effectively connect input signals~\cite{jaegle2021perceiverio,jaegle2021perceiver}. 
Another promising technique is contrastive learning, which enables models to learn similar representations across modalities~\cite{dar2024contrastive,radford2021clip}. 
This approach has demonstrated strong performance in tasks such as radiology report generation when trained on paired X-ray and text samples from the MIMIC-IV dataset~\cite{chambon2022roentgen,johnson2023mimiciv}.
However, contrastive learning requires substantial amounts of data to effectively capture underlying data representations.

Among the multiple modalities especially irregularly sampled time series data (e.g. laboratory measurements in medicine) require special attention.
In a multivariate setting, most methods assume regularly and evenly spaced observational intervals, but these may not necessarily overlap.
Thus, the observations are commonly converted to continuous-time observations with fixed intervals or explicitly modeling the relationship between sensors~\cite{lipton2016timeseries,zhang2022raindrop,zhang2019attain}.
While these approaches have shown good performances, the effort can significantly be reduced by displaying the time series as line graphs and subsequently train a conventional vision transformer (ViT) for classification~\cite{li2023vitst}.
This not only renders the need for additional modeling obsolete but also shows superior performance.
Our approach builds upon the seminal work and extends the setting to the multi-modal case.
We show that the information from multiple modalities can be easily combined without the need for explicit modeling by using conventional vision-text transformer~\cite{liu2023clip,radford2021clip}.
We show that we can integrate a multitude of modalities, more precisely clinical parameters such as e.g. laboratory measurements, data from bedside monitors, and the given medications, electrocardiography scans, and X-ray images by representing each modality as an image and feeding the additional information (e.g. patient metadata) as text to the model.
By displaying each modality as an image (i.e. line graphs) we harmonize the feature extraction pipeline across inputs and acquire multi-modality at no additional cost.
Further, optimally visualizing modalities as images can be interpreted as visual prompt engineering, which demands significantly less effort and coding expertise compared to designing specialized architectures for specific use cases.
This approach lowers barriers to entry, facilitating no- or low-code solutions.

\begin{figure}[t]
    \centering
    \begin{subfigure}{0.45\textwidth}
        \centering
        \includegraphics[height=5.cm]{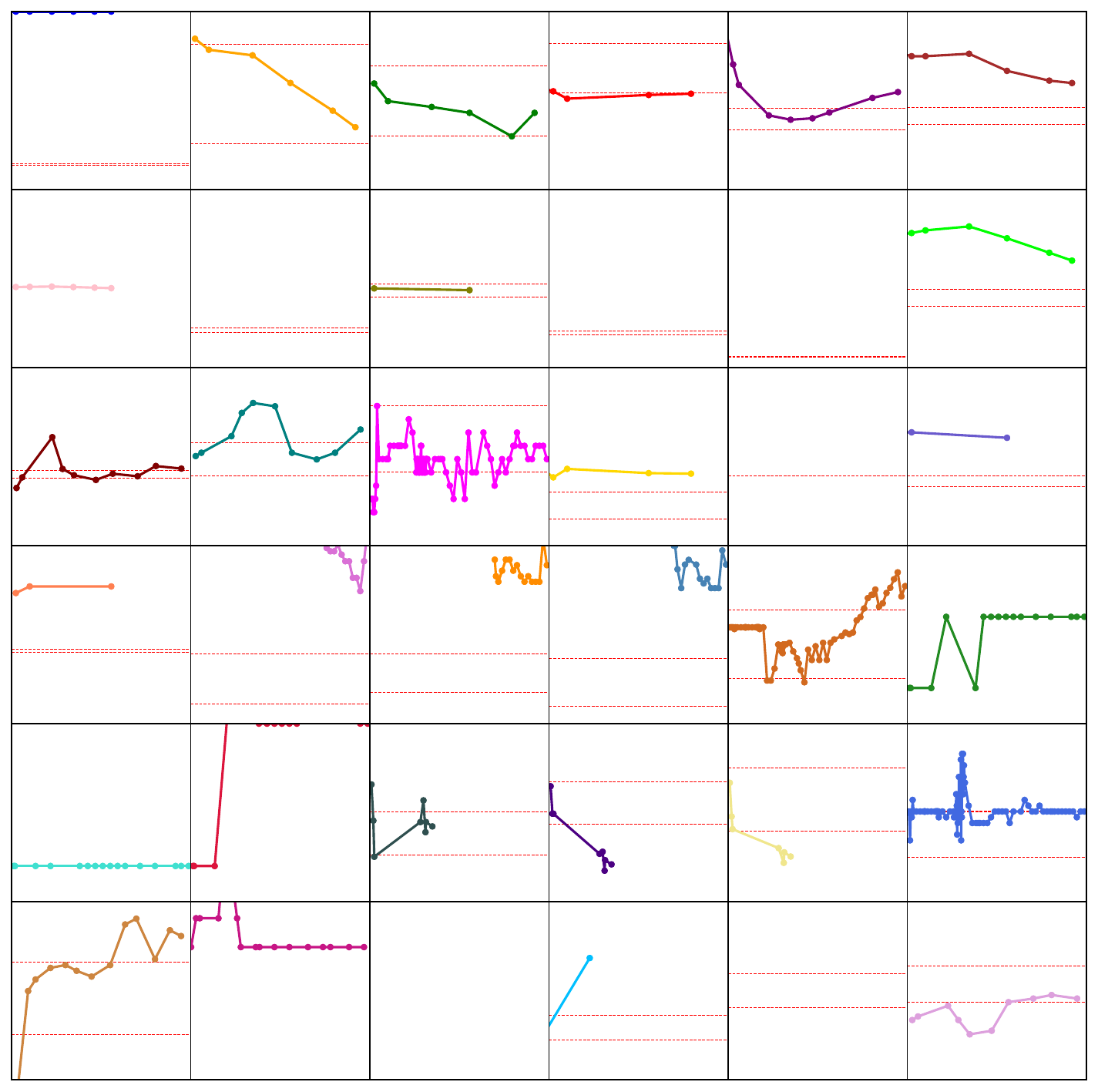}
        \caption{Clinical Measurements (C)}
        \label{fig:lab}
    \end{subfigure}
    \hfill
    \begin{subfigure}{0.45\textwidth}
        \centering
        \includegraphics[height=5.cm,trim=0 10.cm 0 0,clip]{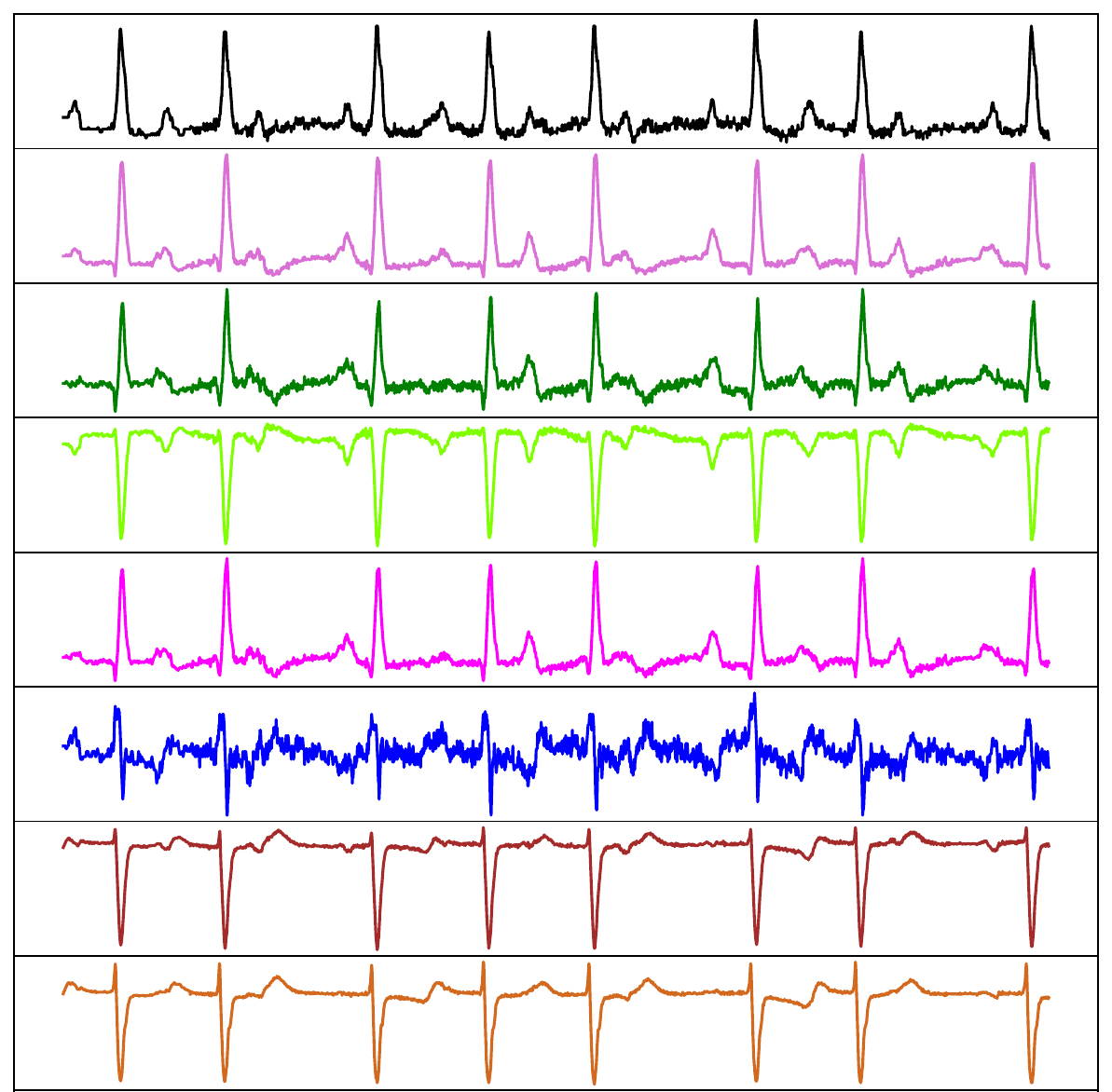}
        \caption{ECG (E)}
        \label{fig:ecg}
    \end{subfigure}
    \begin{subfigure}{0.45\textwidth}
        \centering
        \includegraphics[height=5.cm]{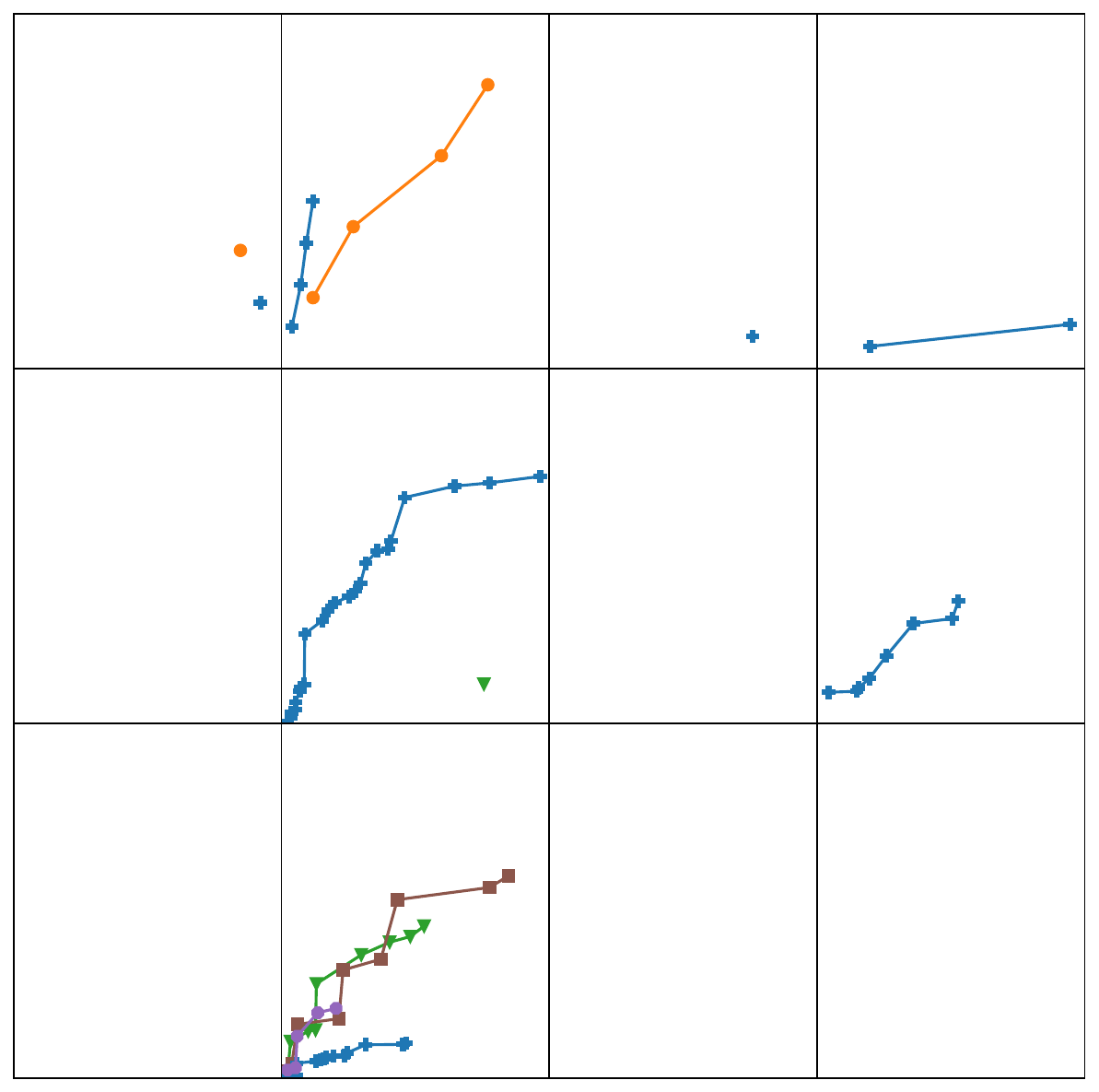}
        \caption{Medications (M)}
        \label{fig:med}
    \end{subfigure}
    \hfill
    \begin{subfigure}{0.45\textwidth}
        \centering
        \includegraphics[height=4.5cm]{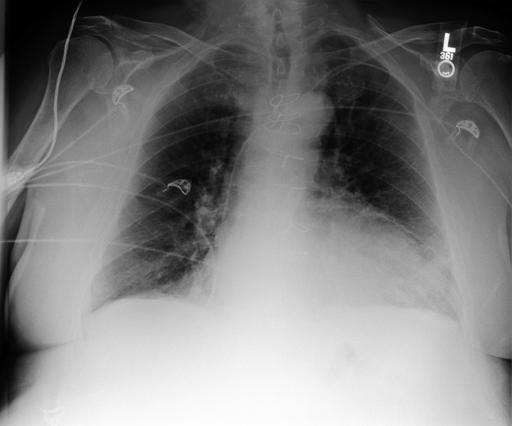}
        \vspace{0.25cm}
        \caption{CXR (X)}
        \label{fig:cxr}
    \end{subfigure}
    \caption{
    The four modalities from the MIMIC-IV dataset used in our model. 
    We tested various formats (e.g., removing "*" markings, preprocessing ECGs), and found the visualized format performed best.
    a) Clinical measurements with x-axis representing 48 hours. b) Medications as cumulative dosage of the different drugs taken. c) 12-lead ECG with minimal preprocessing, from which 8 are shown. d) CXR image. 
    The field names for clinical measurements can be found in Supplementary Figure~\ref{fig:lab-parameters-image} and for medications in Supplementary Figure~\ref{fig:med-parameters-image}. 
    The 12 leads of the ECG are in standard chronological order~\cite{gow2023mimicivecg}.} 
    \label{fig:modalities}
\end{figure}

In the literature the Medical Transformer (MeTra)~\cite{khader2023metra} and medical multi-modal fusion (MedFuse)~\cite{hayat2022medfuse} have been proposed for working with clinical time-series data as well as chest X-ray images from the MIMIC-IV dataset.
MeTra deals with the irregular intervals by taking a day-wise average for each variable and feeding the resulting vector per patient to a linear layer before merging with the extracted features from the X-ray image.
MedFuse discretizes and interpolates the EHR values for harmonized presentation to the model.
They propose to use a long-short term memory (LSTM) network to fuse the extracted features from both modalities, which is particularly useful for the partial instead of paired presence of input modalities per patient.

Compared to the approaches in the literature, the contribution of our work is five-fold:
\begin{itemize}
    \item We are the first to train on more than two modalities from the MIMIC-IV dataset~\cite{johnson2023mimiciv}, including laboratory measurements, medications, electrocardiograms, and X-ray images. Previous multi-modal approaches have been limited by the complexity of modeling more than two modalities.
    \item We significantly reduce this modeling complexity by representing each modality as an image. This approach unifies the data extraction pipeline and facilitates low- to no-code solutions, with the training process primarily influenced by a form of visual prompt engineering.
    \item We outperform two state-of-the-art methods, MeTra~\cite{khader2023metra} and MedFuse~\cite{hayat2022medfuse}, on benchmark tasks such as in-hospital mortality prediction and phenotyping. Our results demonstrate that incorporating more modalities improves model performance.
    \item We enhance the interpretability of our model by visualizing attention maps, which highlight the input regions most influential in the model’s predictions.
    \item We open source our code, model weights, and example scripts, providing accessible tools to apply our method to other datasets with minimal coding effort.
\end{itemize}

\section*{Results}

Not many medical datasets and works exist that provide multi-modal information for patients.
The only publicly available is the Medical Information Mart for Intensive Care (MIMIC) dataset, a deidentified dataset of patients admitted to the emergency department or an intensive care unit at the Beth Israel Deaconess Medical Center in Boston, Massachusetts~\cite{goldberger2000physionet,johnson2023mimiciv}.
It provides detailed information on patient demographics, vital signs, laboratory results, medications, and clinical notes.
Additionally, multiple extensions exists linking the electronic health records to other modalities such as X-ray~\cite{johnson2019mimiccxr} and electrocardiography (ECG)~\cite{gow2023mimicivecg}.

From the MIMIC-IV dataset~\cite{johnson2023mimiciv} two standard benchmarks are extracted~\cite{gupta2022mimicivextract}.
Both task are based on the data from the first 48 hours of a patient's ICU stay.
In the first task \textit{in-hospital mortality} shall be predicted, in the second a collection of 25 diagnosed conditions must be classified (\textit{phenotyping}).
We compare our model, the Vision Transformer for irregular sampled Multi-modal Measurements (ViTiMM), to two approaches from the literature, namely Multimodal Medical Transformer (MeTra)~\cite{khader2023metra} and MedFuse~\cite{hayat2022medfuse}.
However, both methods are only implemented for clinical measurements together with CXR images.
No method exists so far that extends the training task to further modalities such as the administered medications as well as a recorded ECG scan during the ICU stay.
To create maximal compatibility all approaches are only trained on data for which paired clinical measurements and CXR images are available.
The results are summarized in Table~\ref{tab:results}.
The dataset sizes are shown in Table~\ref{tab:dataset_summary} in the methods section.

\subsection*{Uni-modal Training}

First, all methods are evaluated for single modality training.
MeTra and MedFuse are designed to work with only two modalities. 
However, our method easily accommodates training on administered medications and ECG scans by converting their time-series data into visual plots, while utilizing the exact same training procedure as for clinical measurements and CXR images.

\subsubsection*{In-hospital Mortality.}
For clinical measurements (C) our method achieves the highest area under receiver operating curve (AUROC) of 0.837 (MeTra: 0.791, MedFuse: 0.812) and balanced accuracy ($\text{Bal.Acc.}\,=\,(\text{sensitivity}\,+\,\text{specificity})\,/\,2$) of 0.743 (MeTra: 0.609, MedFuse: 0.571).
For area under the precision recall curve (AUPRC) MeTra achieves a value of 0.441, MedFuse 0.448, and ours 0.512.
Classifying CXR images (X) mainly comes down to choosing the better feature extractor.
Here, we can see the superiority of the Swin transformer~\cite{liu2021swin} in our method (AUROC=0.826) compared to the conventional ViT~\cite{dosovitskiy2021vit} in MeTra (0.810) and a ResNet34~\cite{he2016resnet} in MedFuse (0.662).
Training on administered medications (M) and ECG scans (E) yields lower classification accuracy.
For medications our method achieves a AUROC of 0.741 (AUPRC=0.346, Bal.Acc.=0.680), for ECG the AUROC is 0.704 (AUPRC=0.297, Bal.Acc.=0.636).

\subsubsection*{Phenotyping.}
Phenotyping, a multi-label multi-class binary classification task for present diagnoses, is a more challenging task than predicting in-hospital mortality.
However, our method, ViTiMM, outperforms MeTra and MedFuse in the single modality setting for both, clinical measurements (C) and CXR images (X) (see Table~\ref{tab:results}).
By training on C we achieve an AUROC of 0.766 (MeTra: 0.691, MedFuse: 0.705), with training on X the scores are slightly lower: ViTiMM: 0.730, MeTra: 0.667, MedFuse: 0.644.
Again, this is mostly determined by the employed architecture. 

\begin{table}[t]
    \centering
    \caption{
    Results of ViTiMM for the task \textit{In-hospital Mortality} and \textit{Phenotyping} compared to MeTra~\cite{khader2023metra} and MedFuse~\cite{hayat2022medfuse}.
    We compare the three methods for uni-modal training for clinical measurements (C) and X-ray (X) as well as a combination of the two similar to their original publicaiton.
    Extension of both methods to further modalities requires explicit modeling, which must not be done in ViTiMM.
    Thus, by only plotting the other modalities our method can straightforward expand to arbitrary modalities.
    The results per phenotype can be found in Supplementary Table~\ref{tab:results-phenotyping}.
    The corresponding significance tests (pairwise t-test) can be found in Supplementary Table~\ref{tab:p-values-inhospital-mortality}~and~\ref{tab:p-values-phenotyping}.\\
    C: clinical measurements, X: CXR images, M: medications, E: electrocardiography.
    }
    \label{tab:results}
    \begin{tabularx}{\linewidth}{p{0.14\linewidth}p{0.11\linewidth}|C{0.12\linewidth}C{0.12\linewidth}C{0.12\linewidth}|C{0.12\linewidth}C{0.12\linewidth}C{0.12\linewidth}}
        \toprule
        & & \multicolumn{3}{c|}{\textbf{In-hospital Mortality}} & \multicolumn{3}{c}{\textbf{Phenotyping}} \\ 
        Method & Modalities & AUROC & AUPRC & Bal.Acc. & AUROC & AUPRC & Bal.Acc. \\ 
        \midrule
        \multirow{3}{*}{MeTra~\cite{khader2023metra}}
        & C & 0.791 & 0.441 & 0.609 & 0.691 & 0.400 & 0.574 \\
        & X & 0.810 & 0.471 & 0.544 & 0.667 & 0.387 & 0.564 \\
        & $L|X$ & 0.859 & 0.595 & 0.707 & 0.712 & 0.431 & 0.583 \\
        \midrule
        \multirow{3}{*}{MedFuse~\cite{hayat2022medfuse}}
        & C & 0.812 & 0.448 & 0.571 & 0.705 & 0.417 & 0.569 \\
        & X & 0.662 & 0.264 & 0.500 & 0.640 & 0.349 & 0.538 \\
        & $L|X$ & 0.805 & 0.431 & 0.631 & 0.733 & 0.448 & 0.600 \\
        \midrule
        \multirow{6}{*}{\parbox{0.14\linewidth}{ViTiMM \\ (Ours)}} 
        & C & 0.837 & 0.512 & 0.743 & 0.766 & 0.506 & 0.618 \\
        & X & 0.826 & 0.494 & 0.758 & 0.730 & 0.460 & 0.589 \\
        & M & 0.741 & 0.346 & 0.680 & 0.710 & 0.430 & 0.577 \\
        & E & 0.704 & 0.297 & 0.636 & 0.681 & 0.427 & 0.573 \\
        & $C|X$ & \textbf{0.875} & \textbf{0.615} & \textbf{0.776} & \textbf{0.778} & \textbf{0.530} & \textbf{0.636} \\
        & $C|M|X|E$ & \textbf{0.922} & \textbf{0.764} & \textbf{0.847} & \textbf{0.784} & \textbf{0.549} & \textbf{0.659} \\ 
        \bottomrule
    \end{tabularx}
\end{table}

\subsection*{Training on Paired X-ray and Clinical Measurements Data}

Having investigated single modality training we combine two modalities, clinical measurements (C) and CXR images (X) as was done by MedFuse and MeTra as well.

\subsubsection*{In-hospital Mortality.}
The predictive performances of MeTra and ViTiMM can be improved by combining C and X.
When trained on paired clinical measurements and CXR images MeTra achieves an AUROC of 0.859, MedFuse on the other hand only reaches 0.805.
Our method, ViTiMM, outperforms both of them in terms of AUROC of 0.875 but also in AUPRC and balanced accuracy (see Table~\ref{tab:results}).
In their paper Hayat et al. report an AUROC of 0.865 for training on clinical measurements and CXR images~\cite{hayat2022medfuse}.
However, this was achieved when also training on unpaired samples, which we omit to achieve maximal comparibility.
Additionally, our method also outperforms MedFuse trained on partial samples when only trained on paired samples.

\subsubsection*{Phenotyping.}
Similar to \textit{In-hospital mortality} the performance of all methods can be enhanced by using both, clinical measurements and CXR images.
MeTra achieves an AUROC of 0.712, MedFuse of 0.733.
In their paper Hayat et al. again report a slightly higher AUROC of 0.770, which is achieved through training on the partial paired dataset~\cite{hayat2022medfuse}.
However, with our method we again also outperform MedFuse's partial training setting slightly with 0.778.
AUPRC and balanced accuracy follow the same trends.

\subsection*{Extension to All Modalities}

To the best of our knowledge no method in the literature has so far investigated training on the four modalities, clinical measurements (C), medications (M), X-ray images (X), and ECG scans (E). 
Thus, we do not have a direct comparison to our method. 

\subsubsection*{In-hospital Mortality.}
When feeding all modalities, clinical measurements, CXR images, medications, one electrocardiography scan, as well as the corresponding demographics, CXR reports, ECG machine measurements, and patient's diagnoses the predictive performance can significantly be improved.
ViTiMM reaches an AUROC of 0.922 outperforming the two modality training across all methods.
Also in terms of AUPRC (0.764) and balanced accuracy (0.847) training with all modalities represents the benchmark.

\subsubsection*{Phenotyping.}
For \textit{phenotyping} the predictive performance of our method cannot be enhanced to a similar extend as for \textit{In-hospital Mortality}.
Still, we achieve a new benchmark with an AUROC of 0.784 by combining clinical measurements, administered medications, CXR images, and ECG scans.
The AUPRC is 0.549, balanced accuracy lies at 0.659.
In contrast to the mortality prediction task we cannot feed the patient's diagnoses as they shall be predicted from the input data.

\subsection*{Interpretability with Attention Visualization}

\begin{figure}[t]
    \centering
    \begin{subfigure}{0.24\textwidth}
        \centering
        \includegraphics[width=\linewidth]{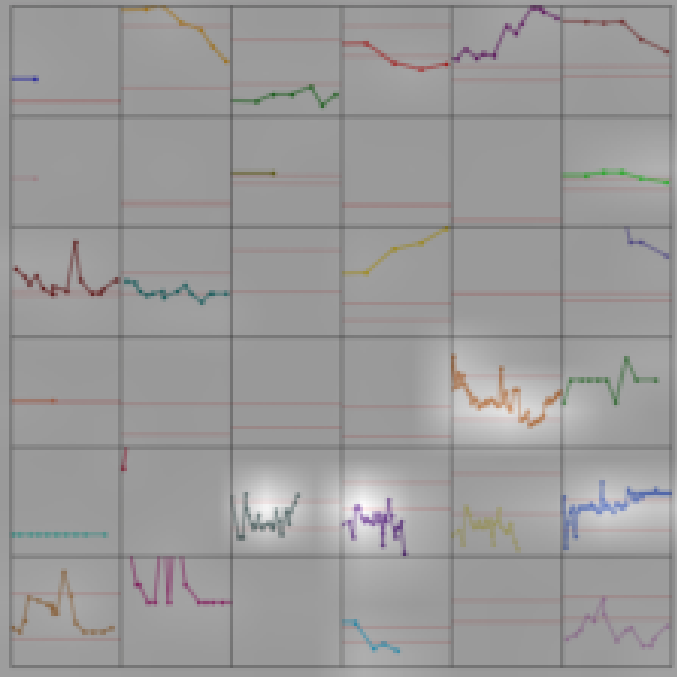}
        \caption{Clinical Measurements}
        \label{fig:attention-lab}
    \end{subfigure}
    \hfill
    \begin{subfigure}{0.24\textwidth}
        \centering
        \includegraphics[width=\linewidth]{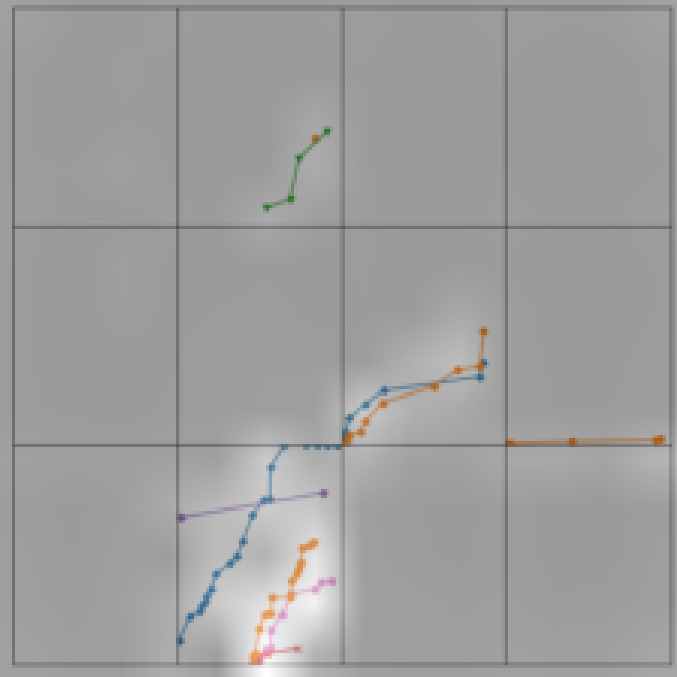}
        \caption{Medications}
        \label{fig:attention-med}
    \end{subfigure}
    \begin{subfigure}{0.24\textwidth}
        \centering
        \includegraphics[width=\linewidth]{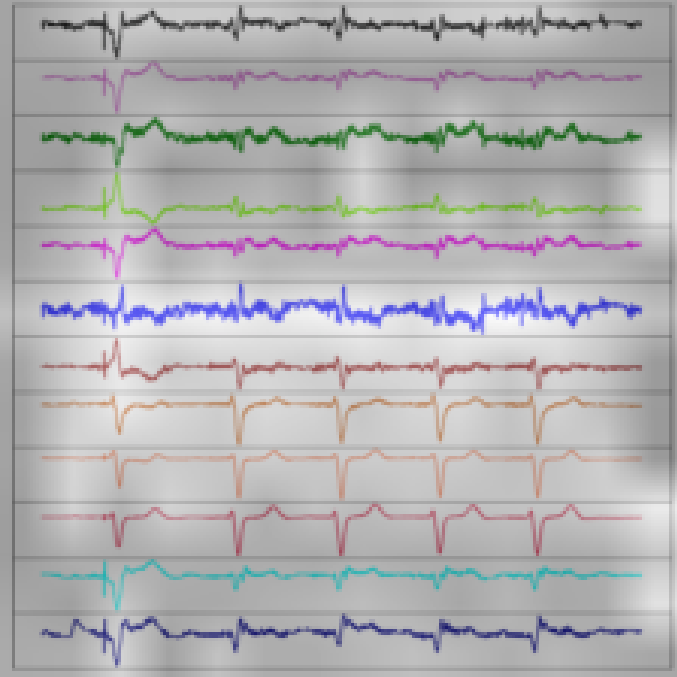}
        \caption{ECG}
        \label{fig:attention-ecg}
    \end{subfigure}
    \hfill
    \begin{subfigure}{0.24\textwidth}
        \centering
        \includegraphics[width=\linewidth]{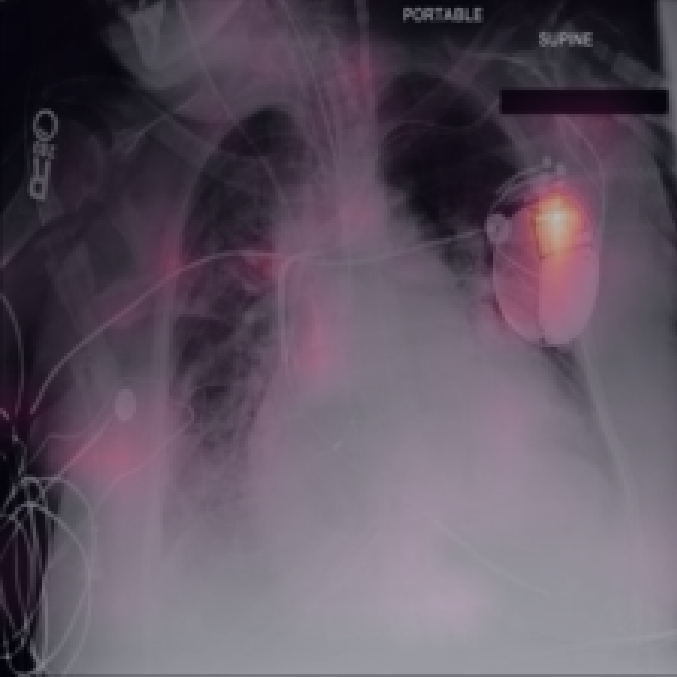}
        \caption{CXR}
        \label{fig:attention-cxr}
    \end{subfigure}
    \begin{subfigure}{0.75\textwidth}
        \centering
        \includegraphics[width=\linewidth, trim=0 2cm 0 0, clip]{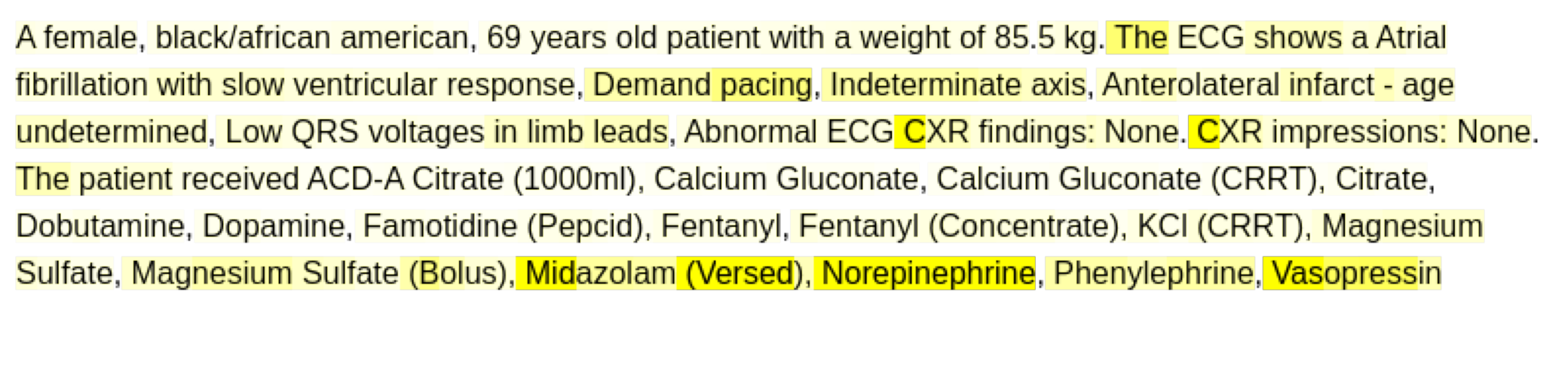} 
        \caption{Textual Metadata}
        \label{fig:attention-text}
    \end{subfigure}
    \caption{
        Interpretability with Visualizing Attention Maps for in-hospital mortality prediction for \texttt{hadm\_id\,=\,30315583}.
        The patient has died within the hospital stay, which the model rightfully predicted.
        The model clearly focuses on specific attributes in the clinical measurements and medications.
        In contrast, the attention for ECG appears more diffuse, suggesting that the model leverages a broader, more global context.
        For the CXR image, the attention is particularly concentrated on the implanted device.
        More examples can be found in the Supplemental Material.
    }
    \label{fig:attention}
\end{figure}

Transformer architectures use the attention mechanism for trading of local and global contexts in vision tasks~\cite{dosovitskiy2021vit}.
When using the conventional ViT the importance of input features can be visualized by overlaying the attention weights of the classification.
The same procedure can be conducted for the text input.
The patient, whose input data is shown in Figure~\ref{fig:attention}, has not survived his hospital stay.
The model specifically focuses on specific features in the clinical measurements data, namely Respiratory Rate (from bottom left row 2, column 6), Non Invasive Blood Pressure Diastolic (2,4), Non Invasive Blood Pressure Diastolic (2,3), and Heart Rate (3,5) (Figure~\ref{fig:attention-lab} with corresponding field names in Supplementary Figure~\ref{fig:lab-parameters-image}).
In the medication inputs the most weight is put on Vasopressors and Inotropes (1,2), more precisely Norephinephrine, Phenylephrine, and Dobutamine, but the model attends to all administered medications to a small degree (Figure~\ref{fig:attention-med} with corresponding field names in Supplementary Figure~\ref{fig:med-parameters-image}).
The attention on the electrocardiography scan is diffuse, the focus rather seems to lie on the overall interaction of all leads.
However, the largest attention lies on lead aVL (blue in Figure~\ref{fig:attention-ecg}).
In the CXR image specific focus lies on the implanted device (Figure~\ref{fig:attention-cxr}).
In the text input the model directs specific attention to the administered medications and the implanted pacemaker.
General attention lies on the patient demographics and ECG findings.
While no findings and impressions are present in the CXR report, the attention peaks in this part indicating general importance if information exists.
Further attention visualizations for both classes in the \textit{In-hospital mortality} task can be found in Supplementary Figures~\ref{fig:attention-mortality-0}~and~\ref{fig:attention-mortality-1}.

\section*{Discussion}

We present a method that can learn from arbitrary modalities by employing vision transformer only by representing the respective data in image format as an extension to ViTST~\cite{li2023vitst}.
We demonstrate that by representing any modality as an image, in a manner broadly analogous to how humans interpret data, multi-modal training can be achieved with minimal effort. 
This is because vision transformers are inherently capable of capturing the context of data in a way that resembles human perception.
This approach simplifies the training process significantly, it is broken down to \textit{visual prompt engineering}, which can be performed with little coding knowledge.
We hope this will lower the barrier of multi-modal training immensely in the future.
Our model not only simplifies multi-modal training in medicine but also shows superior performance on the benchmark tasks of in-hospital mortality prediction and phenotyping~\cite{gupta2022mimicivextract} compared to approaches in the literature~\cite{hayat2022medfuse,khader2023metra}.
To the best of our knowledge we are the first to train one model across the modalities clinical measurements, CXR images, medications, and ECG scans as well as further patient metadata on the MIMIC-IV dataset~\cite{gow2023mimicivecg,johnson2024mimicxr2,johnson2023mimiciv,johnson2019mimiccxr}.

\subsection*{Visual Prompt Engineering Simplifies Modeling of Modalities}

Humans are able to quickly grasp information and trends from visual prepared graphical data.
Vision transformer seem to be able to capture information in a similar manner also observing interactions between different variables.
The approach is especially beneficial for irregularly sampled time series data~\cite{li2023vitst}.
Most approaches need to explicitly model the time dimension by e.g. partitioning the data in fixed time intervals~\cite{lipton16rnn,marlin2012time} or use dedicated architectures such as LSTMs or graph neural networks~\cite{che2018timeseries,mcdermott2023eventstreamgpt,zhang2019attain}.
However, modeling the time dimension might lead to loss of information dependent on e.g. the sampling rate when partitioning observations.
On the other side, when representing the data as line graphs observations can be explicitly marked and missing intervals interpolated in-between.
The model is made aware of the missing information in this area, but is presented a hint for possible values in the interval.
The advantage is emphasized when only comparing the results for clinical measurements (C) across the three methods (MeTra~\cite{khader2023metra}, MedFuse~\cite{hayat2022medfuse}, and ViTiMM) in Table~\ref{tab:results}, where our method obtains the best performance across all metrics.
In MeTra for each clinical parameter in question the mean over the time period is taken; in MedFuse all measurements must be discretized and harmonized to similar time points across variables.

\subsection*{Advantages of the Multi-Modal Approach}

Our results confirm that multi-modal training is beneficial for all investigated methods for both tasks under consideration.
The multi-modal approach enables the extraction of complementary information from the used modalities.
Our approach, ViTiMM, outperforms the two comparative approaches (MeTra~\cite{khader2023metra} and MedFuse~\cite{hayat2022medfuse}) with the same data sources, clinical measurements and CXR images.
We attribute this to potential information loss in the time series data caused by averaging the variables over the respective timeframe (see above).

The key difference between ViTiMM and other approaches in the literature is its straightforward extensibility to other modalities by either representing the data as an image such as e.g. line graphs or providing the data as unstructured text.
Thus, we are able to also feed the medications of each patient despite the high missing ratios (see Supplementary Table~\ref{tab:medications}), the ECG scan, and diagnoses of the patients without altering our encoders.
While our method already achieves the best results compared to MeTra and MedFuse with only clinical measurements and CXR images, the additional modalities lead to even better performances.
Extension of MeTra and MedFuse to other modalities is not straightforward, as each modality must be separately modeled and, thus, receive a different encoder than used currently in their methods.
This would alter the presented methods drastically, which might not align with the intentions of the authors.
In general, especially medications would require much modeling due to high missing ratios and the different dosages of different solutions of the active ingredients.

Still, careful consideration must be made regarding possible confounding variables in the input modalities.
For example, certain of our considered medications, particularly opiods and sedatives, are commonly administered before death.
To assess their impact on the accuracy of predictions, we analyzed the model outputs for those using medications as input (all modalities and medications only). 
We compared the performance across two subgroups: patients who received opiods and/or sedatives (group1) vs. those who did not (group2). 
Comparing AUROC values, we observed a slight increase in performance for group1 (0.942 vs. 0.930 in the all modality setting and 0.768 vs. 0.751 for medications only). 
However, a DeLong test found these differences not statistically significant (p=0.546, p=0.580)~\cite{delong1988delongtest}. 

\subsection*{Benchmark Tasks}

The superiority of using multiple modalities for training due to their complementary information has been proven multiple times~\cite{kline2022multimodalsummary}.
However, public dataset with multiple modalities are scarce.
Especially in medicine the literature focuses on the MIMIC-IV dataset~\cite{hayat2022medfuse,johnson2023mimiciv,khader2023metra,rasekh2024multimodalmimiciv,wang2024multimodalmimiciv}.
The MIMIC-IV is a role model for an open source multi-modal dataset covering several modalities namely clinical measurements, medications, CXR images, ECG scans, and recently also echocardiography (ECHO) data~\cite{gow2023mimicivecho}.
Benchmark tasks are defined with a extraction pipeline establishing comparability between studies~\cite{gupta2022mimicivextract}.
However, some discrepancies still occur when paired or partial samples across modalities are used~\cite{hayat2022medfuse,khader2023metra}.
Further public multi-modal benchmark datasets are needed to examine the generalizability of our and other methods.
In future work we also plan to incorporate echocardiography data into our pipeline, which technically is straightforward as they are naturally represented as images.
However, echocardiography data has a very large intra-patient variance due to multiple views, which are not explicitly labeled or identified in the dataset.


\section*{Methods}

This manuscript's study and results adhere to all pertinent ethical guidelines and uphold ethical standards in both research conduct and manuscript preparation, in accordance with all relevant laws and regulations concerning human subject treatment. 
All models were trained on publicly available datasets described below and tested for their performance in predicting the survival of patients in intensive care.

\begin{figure}[t]
    \centering
    \includegraphics[width=0.7\textwidth]{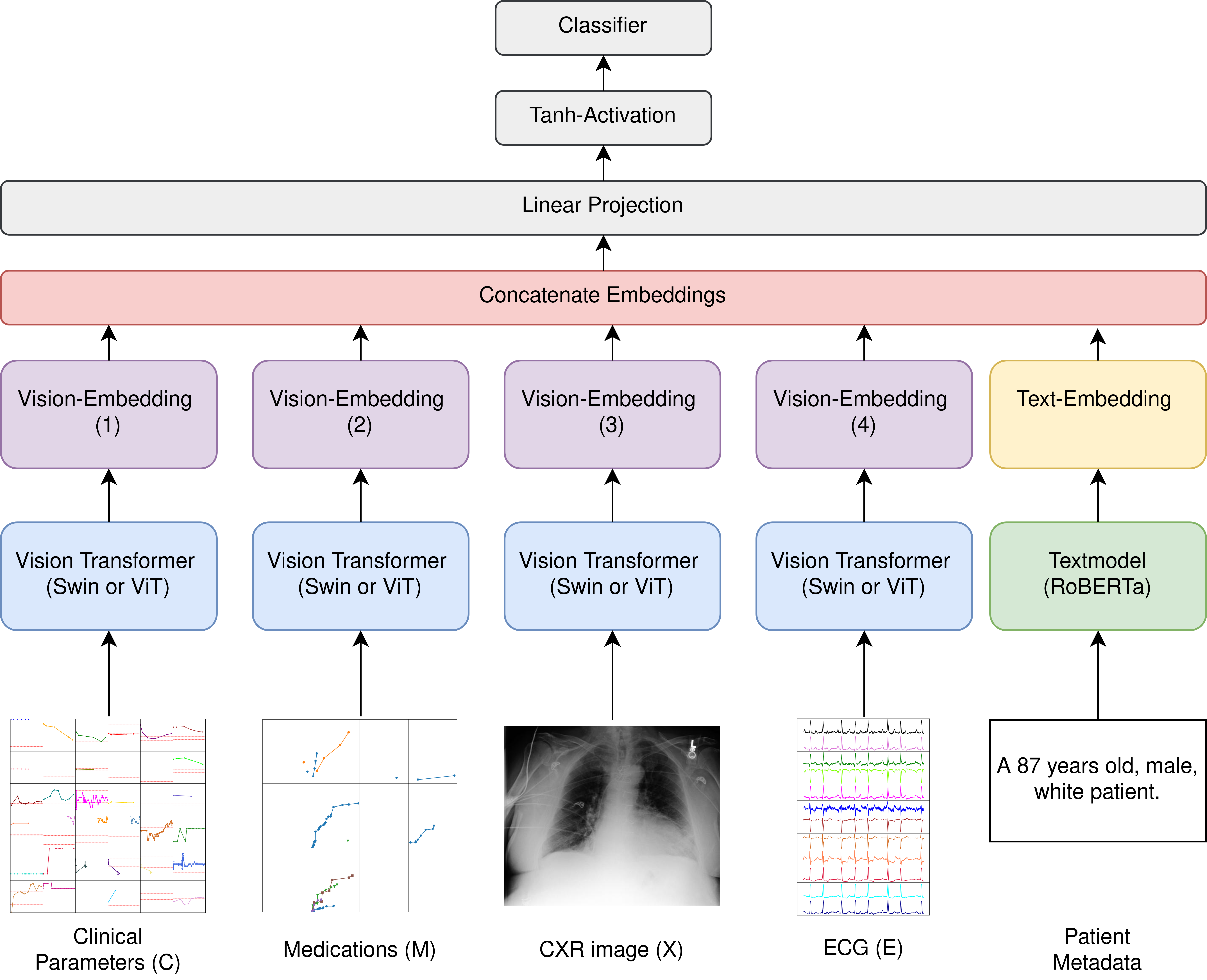}
    \caption{
    Multi-modal medical vision transformer with arbitrary modalities as images.
    Each modality is represented as an image i.e. clinical measurements, administered medications, and electrocardiography scans are displayed as line graphs.
    No changes must be made to CXR images.
    Each image can be fed into a vision transformer (conventional ViT~\cite{dosovitskiy2021vit} or SWIN~\cite{liu2021swin}).
    Additionally, patient metadata is fed as text into a RoBERTA model~\cite{liu2019roberta}.
    As embeddings are added after feature extraction, arbitrary modalities can be added or subtracted.
    }
    \label{fig:method}
\end{figure}

\subsection*{Data and Benchmarks}

The most frequently used dataset with matched time-resolved samples on a patient level across multiple modalities is the the Medical Information Mart for Intensive Care (MIMIC)~\cite{johnson2023mimiciv}.
MIMIC-IV is a freely accessible dataset that contains extensive clinical information from de-identified patients admitted to the emergency department or an intensive care unit at Beth Israel Deaconess Medical Center (BIDMC) in Boston, Massachusetts, between 2008 and
2019.
It consists of two main modules. 
The hospital-wide EHR module includes patient and admission information, medication administrations, billed diagnoses, microbiological data, laboratory measurements and hospital performance-related information. 
The ICU-specific module comprises detailed records to ICU stays, including intravenous and fluid administrations, patient excretions, ventilation data, and clinical interventions.
In total, the dataset contains 299,712 unique patient identifiers, 431,231 hospitalizations and 73,181 intensive care unit (ICU) stays.
All data is publicly available via Physionet~\cite{goldberger2000physionet}.

For subsets of the patients contained in the MIMIC-IV dataset further modalities are published, more precisely chest X-ray (CXR) images (MIMIC-CXR)~\cite{johnson2024mimicxr2,johnson2019mimiccxr} as well as electrocardiography scans (MIMIC-IV-ECG)~\cite{gow2023mimicivecg}.
The MIMIC-CXR dataset includes 377,110 CXR images (frontal and lateral views) corresponding to 227,835 radiological examinations collected between 2011 and 2016. 
The radiographs are in DICOM format and sourced from the hospital's Picture Archiving and Communication System (PACS).
A free-text radiology report is assigned to each examination.
MIMIC-IV-ECG provides electrocardiograms from 161,352 different patients and includes 800,035 diagnostic ECGs recorded between 2008 and 2019.
While 55\% can be linked to a hospitalization record and 25\% to a visit to the emergency department, for 20\% no link to the MIMIC-IV dataset can be established.

To provide rigorous evaluation to other methods, benchmark tasks need to be defined, of which two exist that were also used in other works~\cite{gupta2022mimicivextract,hayat2022medfuse,khader2023metra}.
\begin{itemize}
    \item \textbf{In-hospital mortality}: This binary classification task aims to predict in-hospital mortality based on the first 48 hours of ICU stay.
    Patients with ICU stays shorter than 48 hours are excluded.
    \item \textbf{Phenotype classification}: In this multi-label classification task a set of 25 chronic, mixed, and acute care conditions are assigned to a patient in an ICU stay (e.g. acute myocardial infarction or shock).
\end{itemize}
In both tasks the instance is paired with the last CXR image and ECG scan during the first 48 hours of the ICU stay.
Table~\ref{tab:dataset_summary} shows the resulting dataset splits with their respective quantities.
The age of the patients ranges from 18 to 91 years with an average of $64 \pm 16$\,years.
Approximately 55\% of patients were male.
To enable to comparison to MeTra we only use samples that have paired EHR and CXR data.
Still, not all of these patients have an ECG scan, which highlights the possible extensibility of our method to unpaired samples.

\begin{table}[t]
    \centering
    \caption{Summary of matched MIMIC-IV, -CXR, and -ECG datasets and the corresponding splits. The splits were stratified based on in-hospital mortality. The phenotyping labels are present for all patients, their individual prevalence can be found in Supplementary Table~\ref{tab:results-phenotyping}.}
    \begin{tabularx}{\textwidth}{p{0.32\linewidth}C{0.17\linewidth}C{0.17\linewidth}C{0.17\linewidth}C{0.17\linewidth}}
        \toprule
        Variable & All & Train Set & Validation Set & Test Set \\
        \midrule
        Patients & 6125 & 4396 & 472 & 1257 \\
        Unique Stay IDs & 6798 & 4885 & 540 & 1373 \\
        CXR Images & 6798 (100\%) & 4885 (100\%) & 540 (100\%) & 1373 (100\%) \\
        ECG Scans & 3840 (56\%) & 2753 (56\%) & 301 (56\%) & 786 (57\%) \\
        Clinical Parameters & 6798 (100\%) & 4885 (100\%) & 540 (100\%) & 1373 (100\%) \\
        In-hospital mortality & 1002 (16\%) & 717 (16\%) & 76 (16\%) & 209 (17\%) \\
        \bottomrule
    \end{tabularx}
    \label{tab:dataset_summary}
\end{table}

\subsection*{Model Input Transformations}

To enable training on time series data with vision transformers a transformation into images more precisely line graphs is necessary.
All three time series modalities need different preprocessing to be afterwards processed in a similar way.
Only X-ray scans need no further preprocessing as they are already in image format (Figure~\ref{fig:cxr}).

\subsubsection*{Irregularly sampled Time Series Line Graphs.}

To visualize the course of temporal data points most often line graphs are used, where each observation is marked by its time (x-axis) and value (y-axis).
These observations are connected with straight lines in chronological order, which accounts for interpolating values in between.
This approach is agnostic to the type of time series under consideration and generalizes as we show for different modalities.
Essentially, this is a form of visual prompt engineering similar to language models, where users refine and optimize natural language prompts to improve model performance~\cite{li2023vitst}.
To distinguish measurements from interpolation we use a "*" symbol to indicate observations.
We plot each variable in a separate plot with a distinct color for better differentiation as was found beneficial for model training (see Figure~\ref{fig:lab})~\cite{li2023vitst}.
We omit tick labels as well as other graphical components as an observation's position already signals its relative time and magnitude.

\subsubsection*{Clinical Measurements.}

We identified 36 different clinical variables in consultation with expert physicians (Supplementary Table~\ref{tab:clinical-parameters}).
Before plotting, each variable is standardized to zero mean and unit variance with $(x-\mu)/\sigma$.
Subsequently, each variable was displayed in its own quadrant with a distinct color.
If no measurements were present for a specific patient the quadrant is left blank.
Supplementary Table~\ref{tab:clinical-parameters} shows the missing ratios for all parameters.
Each variable is consistently visualized in the same quadrant and color for all patients. 
An example in shown in Figure~\ref{fig:lab}, the corresponding variable names can be found in Supplementary Figure~\ref{fig:lab-parameters-image}.
For providing further guidance to the model for each variable the normal lower and upper range are visualized as red lines in each quadrant.

\subsubsection*{Medications.}

Compared to clinical measurements, visualizing medications required more effort.
Different prescriptions could be given in different solutions which were encoded in different drug names.
So, we first grouped the medications into ten categories such as e.g. beta-blockers or antiarrhythmics and assigned different substances into the categories (Supplementary Table~\ref{tab:medications}).
Each category received one quadrant in the final block, whereas each substance within a category was a separate line graph with a consistent color across all patients (Figure~\ref{fig:med} and Supplementary Figure~\ref{fig:med-parameters-image}).
For each substance the cumulative dosage is shown, i.e. each line is consistently increasing over time.
Similar to clinical measurements to remove outliers we clip the cumulative values at the 95th quartile.
Afterwards all cumulative amounts are normalized between 0 and 1 with the maximum amount across all patients.
Due to the high number of different administered drugs (e.g. different solution combinations) each medication category exhibits high missing ratios.

\subsubsection*{Electrocardiography Scans.}

We sampled the waveforms at 500 Hz and performed standardization~\cite{mckeen2024ecgfm}.
We extracted a 5 s from the signal due to limited input resolution.
Since ECGs are regularly sampled a visualization in line graphs is straightforward (Figure~\ref{fig:ecg}).
To preserve important details each of the twelve leads is plotted in a new row.
As for clinical measurements and medications we also resort to assigning a specific color to each lead for better differentiation.

\subsubsection*{Patient Metadata.}

To enable learning on categorical patient metadata we feed important information as text to our language model~\cite{liu2019roberta}.
These include the patient's sex, age, and ethnicity as well as insurance status.
Further, we feed the findings and impressions from each CXR report, the machine measurements i.e. diagnoses from the ECG such as e.g. myocardial infarction or bundle branch blocks, and the diagnoses encoded in the international classification of diseases (ICD) codes.
Last, we also feed the prescribed medications as string without the administered dosages as we found this improves predictive performance.

\subsection*{Architecture with Vision Transformers}

The line graphs involve both local (the temporal course of a single variable) and global (the correlation among several variables) contexts.
Given enough data vision transformers have been shown to excel at maintaining spatial information and stronger capabilities to capture both, local and global contexts~\cite{dosovitskiy2021vit,liu2021swin,toelle2024fedkd}.

The conventional vision transformer (ViT) splits the image into fixed size patches that are subsequently linearly projected and extended with position embeddings~\cite{dosovitskiy2021vit}.
The resulting vectors together with a dedicated classification token are fed into a multi-head self attention modules.
Global context is modeled with attention between all pairs of patches, which can become computationally expensive.
Swin (shifted window) transformer on the other hand trades of local and global inter-unit interactions by formulating a hierarchical representation~\cite{li2023vitst}.
Small-sized patches in earlier layers and gradually merged with neighboring ones deeper in the network.
Self-attention is computed within each non-overlapping window, which allows for capturing local intra-variable interactions and temporal dynamics of a single line graph i.e. one variable~\cite{li2023vitst}.
By shifting the window for creating patches in the subsequent layer and merging the attention connection between different windows the capture of global context i.e. interactions between different variables is enabled.

For each modality we use a separate Swin transformer~\cite{liu2021swin} pretrained on ImageNet~\cite{deng2009imagenet}.
The patient metadata is fed into a RoBERTa text model ~\cite{liu2019roberta} similar to Li et al.~\cite{li2023vitst}.
The extracted features are concatenated before being linearly projected and fed into a final classification layer (Figure~\ref{fig:method}).
While early fusion approaches are also possible, we deliberately chose late fusion as this allows for seamless addition and subtraction of modalities by only changing the number of input features in the linear projection layer.
Early fusion on the other hand would require architectural changes within the feature extractors.
Unless otherwise specified we use an image size of $384 \times 384$ with a channel-wise normalization taken from ImageNet~\cite{deng2009imagenet}.
For the text model, we chose a Byte-Pair Encoding (BPE) tokenizer with a context length of 512 tokens, consistent with the input requirements of the RoBERTa architecture.

\subsection*{Baseline Comparisons}

To enable rigorous evaluation of our method, we compare against two baseline methods from the literature that attempted to solve the same tasks, in-hospital mortality prediction and phenotyping on the MIMIC-IV dataset~\cite{johnson2023mimiciv}, name MedFuse~\cite{hayat2022medfuse} and MeTra~\cite{khader2023metra}.
Both methods work on (paired) CXR images and electronic health records (EHR) i.e. clinical measurements.
In MeTra, the CXR images are processed by a conventional ViT that converts the images into latent representations (tokens).
For each of the clinical measurements a mean over 48 hours is taken, if a parameter is missing it is filled with the overall mean.
The resulting vector is projected with a linear layer.
Both representations are concatenated and fed into four multi-head self attention blocks~\cite{vaswani2017attention}, before being processed by a classification layer.
MedFuse processes the CXR images with a ResNet34 encoder~\cite{he2016resnet}.
The time series data from the clinical measurements is fed to a two-layer long short-term memory (LSTM) network~\cite{hochreiter1997lstm}.
To also enable learning on unpaired data, one or both representation is fed sequentially into a one-layer LSTM.

\section*{Data Availability}

All data is publicly available from the MIMIC database~\cite{goldberger2000physionet,johnson2023mimiciv} on PhysioNet (MIMICIV: \url{https://physionet.org/content/mimiciv/3.1}, MIMIC-CXR-JPG: \url{https://physionet.org/content/mimic-cxr-jpg/2.0.0}, and MIMIC-IV-ECG: \url{https://physionet.org/content/mimic-iv-ecg/1.0}). The code to extract the chest radiographs and corresponding clinical parameters can be found in the GitHub repository linked in the code availability section.

\section*{Code Availability}

Following the FAIR criteria (findability, accessibility, interoperability, and reusability) in scientific research all code used in this study will be made publicly available.
The code for cohort creation, data preprocessing i.e. plot generation, training, and evaluation will be made open source available. 

\section*{Acknowledgements}

The project was funded by the Multidimension AI Project from the Carl-Zeiss Foundation (P2022-08-010).

\section*{Author Contributions}

MT and SE developed the idea for the study.
MT and MS developed the presented method, conducted the experiments and evaluations, and wrote the manuscript.
SE and PW initiated the overall project and significantly helped to shape the methods and the manuscript. 
SF and CR provided medical expertise for choosing important input variables as well as interpreting the results and helped with revising the final manuscript.
SZ, CD, BM, NF, PW, and SE provided guidance and helped with revising the final manuscript.

\section*{Competing interests}

NF reports speaker honoraria, presentations or advisory board consultations from AstraZeneca, Bayer AG, Boehringer Ingelheim, Novartis, Pfizer, Daiichi Sankyo Deutschland.
SE reports speaker honorarium from Boehringer Ingelheim.
None are related to the content of the manuscript.
The other authors declare no conflicts of interest.

\bibliographystyle{splncs04}
\bibliography{Paper}

\newpage

\FloatBarrier
\section*{Supplementary Information}
\FloatBarrier

\renewcommand{\figurename}{Supplementary Figure}
\renewcommand{\tablename}{Supplementary Table}
\setcounter{figure}{0}
\setcounter{table}{0}


\FloatBarrier
\subsection*{Hyperparameter}
\FloatBarrier

\begin{table}[H]
    \centering
    \caption{
    Hyperparameter of ViTiMM.
    For training more than one modality the batch size had to be reduced to 4.
    Training was performed on two Nvidia RTX 3090 Ti.
    3 epochs were sufficient for model training, afterwards slight overfitting could be observed similar to Li et al.~\cite{li2023vitst}.
    }
    \label{tab:hyperparameter}
    \begin{tabularx}{\linewidth}{p{0.19\linewidth}C{0.1\linewidth}C{0.11\linewidth}C{0.07\linewidth}C{0.1\linewidth}C{0.18\linewidth}C{0.08\linewidth}C{0.11\linewidth}}
        \toprule
        \textbf{Experiment} & \textbf{Model} & \textbf{Image Size} & \textbf{Epochs} & \textbf{Optimizer} & \textbf{LRs} & \textbf{$L_2$-Reg.} & \textbf{Batch Size} \\
        \midrule
         In-hospital Mortality & Swin-Large & 384 & 3 & AdamW & {\scriptsize $10^{-5}, 5 \times 10^{-6}, 10^{-6}$} & $3e-8$ & 8 (4) \\
         \midrule
         Phenotyping & Swin-Large & 384 & 3 & AdamW & $10^{-5}$ & $3e-8$ & 8 (4) \\
         \bottomrule
    \end{tabularx}
\end{table}

\FloatBarrier
\subsection*{Input Parameter: Clinical Measurements and Medications}
\FloatBarrier

\begin{figure}[H]
    \centering
    \includegraphics[width=0.7\linewidth]{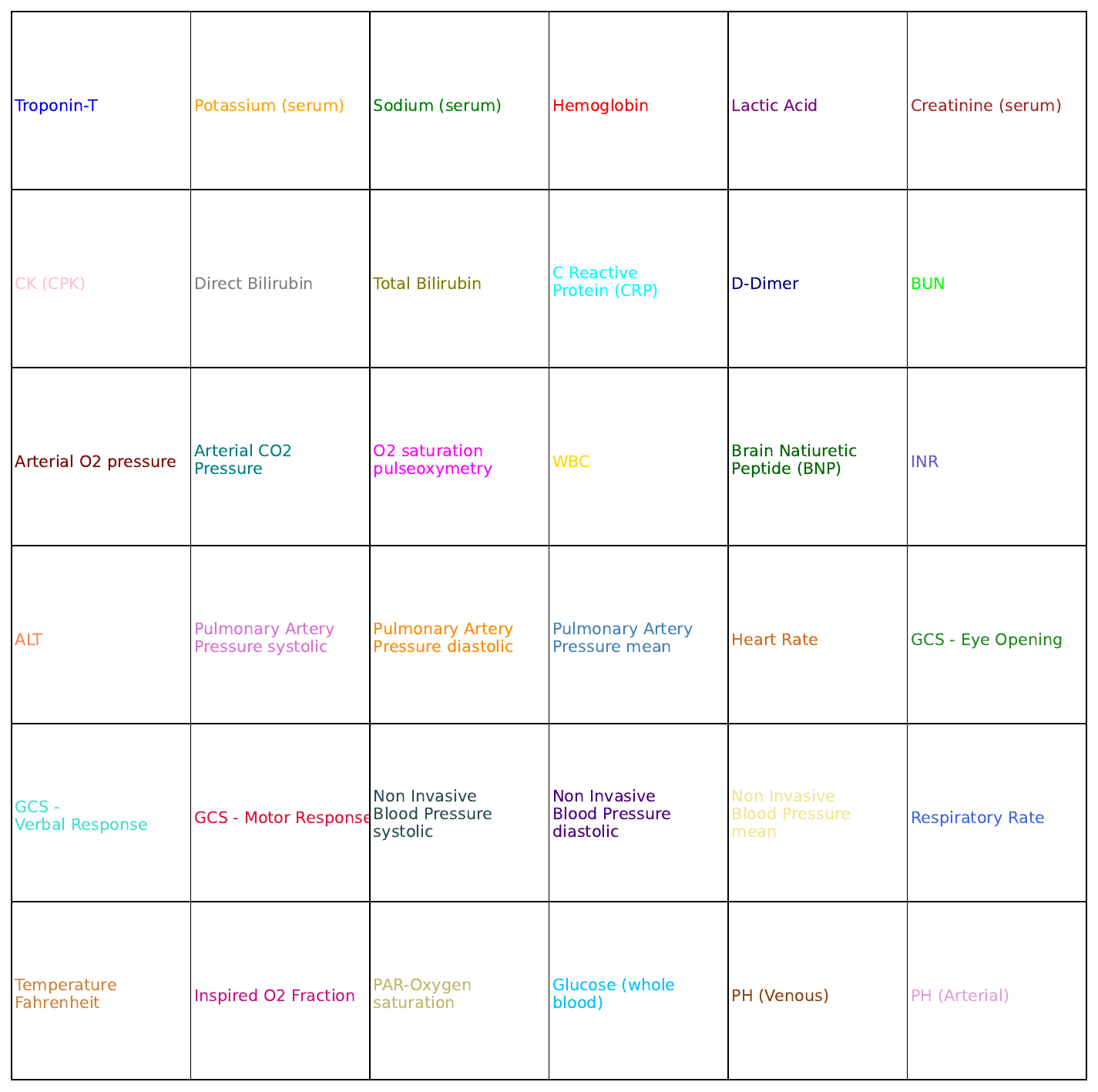}
    \caption{All clinical measurements (C) in their respective field color in the line plots.}
    \label{fig:lab-parameters-image}
\end{figure}

\begin{figure}
    \centering
    \includegraphics[width=0.8\linewidth]{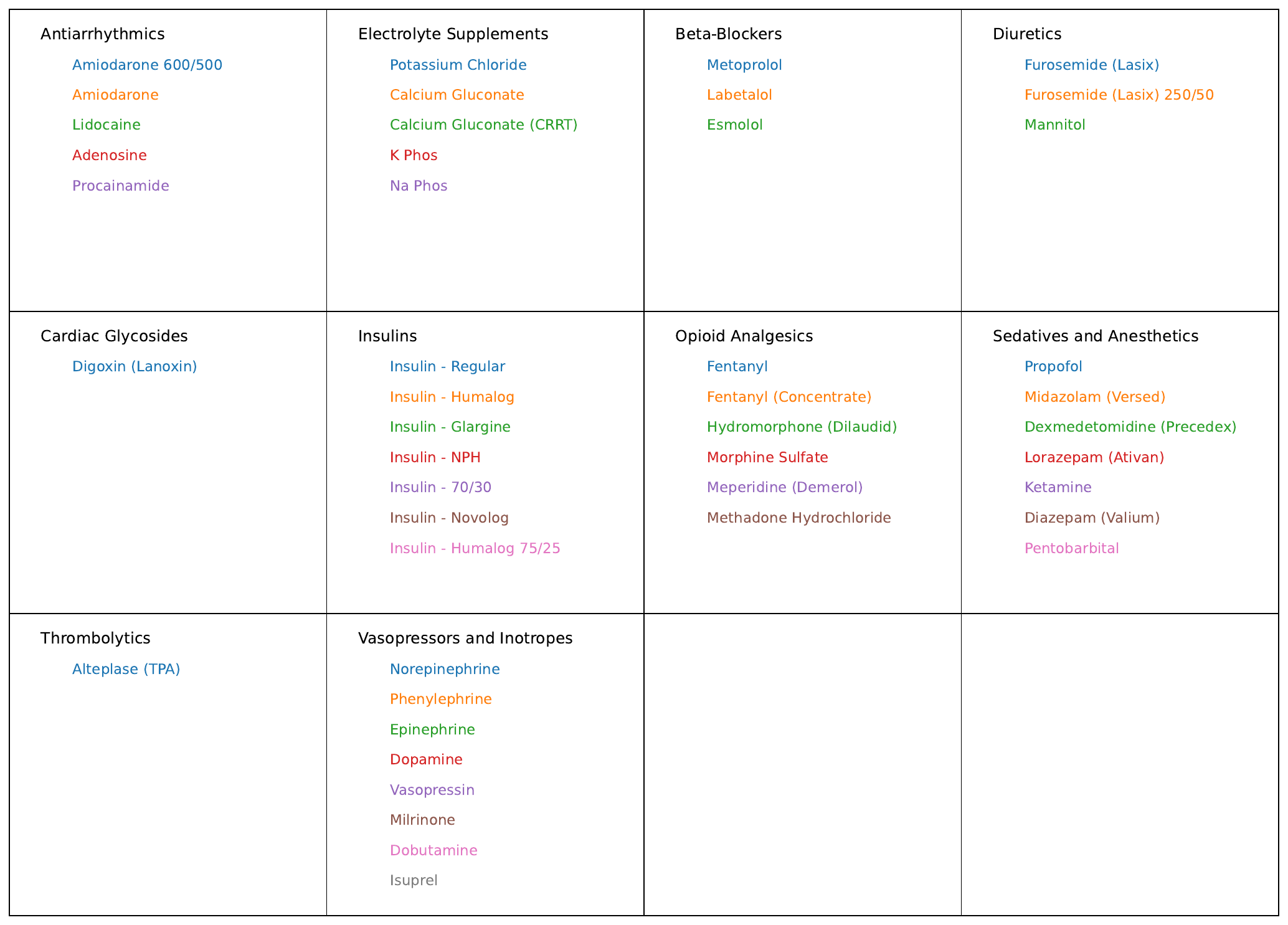}
    \caption{All medications (M) in their respective field color in the line plots.}
    \label{fig:med-parameters-image}
\end{figure}

\begin{table}[t]
\centering
    \caption{
    Summary of Clinical Parameters with Statistics, Unit, and Missing Data.
    CK (CPK): Creatine Kinase, CRP: C Reactive Protein, WBC: White Blood Cells, BNP: Brain Natriuretic Peptide, INR: International Normalized Ratio, ALT: Alanine Aminotransferase, SYS: systolic, DIA: diastolic, PAP: Pulmonary Artery Pressure, NIBP: Non Invasive Blood Pressure, PAR: partial.
    }
    \label{tab:clinical-parameters}
    \begin{tabularx}{\linewidth}{p{0.2\linewidth}C{0.12\linewidth}C{0.12\linewidth}C{0.12\linewidth}C{0.12\linewidth}C{0.15\linewidth}C{0.15\linewidth}}
        \toprule
        \textbf{Variable} & \textbf{Mean} & \textbf{Std} & \textbf{Lower} & \textbf{Upper} & \textbf{Unit} & \textbf{Missing (\%)} \\
        \midrule
        Troponin-T & 0.70    & 1.34     & 0.00   & 0.01   & ng/mL       & 47.18 \\
        Potassium (serum) & 4.08    & 0.54     & 3.30   & 5.10   & mmol/L      & 0.01 \\
        Sodium (serum) & 139.09  & 5.05     & 133.00 & 145.00 & mmol/L      & 0.01  \\
        Hemoglobin & 9.90    & 1.93     & 12.00  & 16.00  & g/dL (M)    & 0.01  \\
        Lactic Acid & 2.52    & 2.04     & 0.50   & 2.00   & mmol/L      & 14.89 \\
        Creatinine (serum) & 1.38    & 1.22     & 0.40   & 1.10   & mg/dL (M)   & 0.01  \\
        CK (CPK) & 1768.03 & 4669.05  & NaN    & NaN    & U/L         & 50.25 \\
        Direct Bilirubin & 3.55    & 3.93     & 0.00   & 0.30   & mg/dL       & 86.33 \\
        Total Bilirubin & 2.12    & 3.38     & 0.00   & 1.50   & mg/dL       & 29.20 \\
        CRP & 73.18   & 77.64    & 0.00   & 5.00   & mg/L        & 94.48 \\
        D-Dimer & 8042.00 & 6100.88  & 0.00   & 0.50   & µg/mL       & 97.69 \\
        BUN & 28.31   & 22.58    & 7.00   & 20.00  & mg/dL       & 0.01  \\
        Arterial O2 pressure & 143.81  & 81.18    & 85.00  & 105.00 & mmHg        & 20.95 \\
        Arterial CO2 Pressure & 40.85   & 8.94     & 35.00  & 45.00  & mmHg       & 20.95 \\
        O2 pulseoxymetry & 96.19   & 2.24     & 95.00  & 100.00 & \%          & 0.00  \\
        WBC & 11.71   & 5.59     & 0.00   & 5.00   & cells/µL    & 0.01  \\
        BNP & 6063.74 & 9257.13  & 0.00   & 100.00 & pg/mL       & 89.44 \\
        INR & 1.46    & 0.54     & 0.90   & 1.10   & -           & 3.46  \\
        ALT & 203.73  & 669.92   & 0.00   & 40.00  & U/L         & 28.99 \\
        PAP-SYS & 35.89   & 8.94     & 15.00  & 30.00  & mmHg        & 86.44 \\
        PAP-DIA & 18.44   & 5.37     & 8.00   & 15.00  & mmHg        & 86.44 \\
        PAP-MEAN & 25.13   & 6.30     & 10.00  & 20.00  & mmHg        & 86.23 \\
        Heart Rate & 85.75   & 17.27    & 60.00  & 100.00 & beats/min   & 0.00  \\
        GCS-Eye Opening & -     & -      & 1.00   & 4.00   & -           & 0.00  \\
        GCS-Verbal Response & -     & -      & 1.00   & 5.00   & -           & 0.00  \\
        GCS-Motor Response & -     & -      & 1.00   & 6.00   & -           & 0.00  \\
        NIBP-SYS & 119.52  & 20.51    & 90.00  & 120.00 & mmHg        & 0.34  \\
        NIBP-DIA & 65.51   & 14.06    & 60.00  & 80.00  & mmHg        & 0.34  \\
        NIBP-MEAN & 78.85   & 14.09    & 70.00  & 100.00 & mmHg        & 0.34  \\
        Respiratory Rate & 19.80   & 5.14     & 12.00  & 20.00  & breaths/min & 0.00  \\
        Temperature & 98.51   & 1.02     & 97.00  & 99.50  & °F          & 0.49  \\
        Inspired O2 Fraction & 44.90   & 10.28    & 0.21   & 1.00   & -           & 17.73 \\
        PAR-O2 saturation & -     & -      & 95.00  & 100.00 & \%          & 82.69 \\
        Glucose (whole blood) & 139.71  & 42.06    & 70.00  & 105.00 & mg/dL      & 49.85 \\
        PH (Venous)  & 7.36    & 0.09     & 7.31   & 7.41   & -           & 47.69 \\
        PH (Arterial) & 7.38    & 0.08     & 7.35   & 7.45   & -           & 20.55 \\
        \bottomrule
    \end{tabularx}
\end{table}

\begin{table}[t]
    \centering
    \caption{Medication Summary with Categories, Statistics, and Missing Values}
    \label{tab:medications}
    \begin{tabularx}{\linewidth}{p{0.2\linewidth}C{0.3\linewidth}C{0.1\linewidth}C{0.1\linewidth}C{0.1\linewidth}C{0.2\linewidth}}
        \toprule
        \textbf{Category} & \textbf{Medication} & \textbf{Mean} & \textbf{Std} & \textbf{Unit} & \textbf{Missing (\%)} \\
        \midrule
        \multirow{5}{*}{Antiarrhythmics} 
        & Amiodarone 600/500        & 320.05  & 164.60  & mg    & 95.72 \\
        & Amiodarone                & 165.50  & 76.72   & mg    & 96.47 \\
        & Lidocaine                 & 541.49  & 574.92  & mg    & 99.63 \\
        & Adenosine                 & 6.93    & 2.61    & mg    & 99.62 \\
        & Procainamide              & 656.89  & 312.71  & mg    & 99.97 \\
        \hline
        \multirow{5}{*}{Electrolyte Supplements}        
        & Potassium Chloride        & 18.25   & 20.57   & mEq   & 59.56 \\
        & Calcium Gluconate         & 2.04    & 1.40    & g     & 63.90 \\
        & Calcium Gluconate (CRRT)  & 17.01   & 50.70   & g     & 97.98 \\
        & K Phos                    & 19.23   & 10.72   & mmol  & 94.47 \\
        & Na Phos                   & 19.82   & 10.71   & mmol  & 95.28 \\
        \hline
        \multirow{3}{*}{Beta-Blockers} 
        & Metoprolol                & 5.14    & 2.01    & mg    & 85.42 \\
        & Labetalol                 & 46.41   & 116.89  & mg    & 93.60 \\
        & Esmolol                   & 848.74  & 769.93  & mg    & 98.63 \\
        \hline
        \multirow{3}{*}{Diuretics} 
        & Furosemide (Lasix)        & 42.81   & 39.13   & mg    & 67.87 \\
        & Furosemide (Lasix) 250/50 & 114.20  & 103.18  & mg    & 95.47 \\
        & Mannitol                  & 46.39   & 26.35   & g & 98.81 \\
        \hline
        Cardiac Glycosides & Digoxin (Lanoxin)         & 14.69   & 49.34   & mg    & 99.16 \\
        \hline
        \multirow{7}{*}{Insulins} 
        & Insulin - Regular         & 7.86    & 16.23   & units & 69.96 \\
        & Insulin - Humalog         & 4.57    & 4.37    & units & 81.54 \\
        & Insulin - Glargine        & 21.14   & 15.04   & units & 86.51 \\
        & Insulin - NPH             & 18.15   & 13.08   & units & 98.98 \\
        & Insulin - 70/30           & 22.26   & 12.90   & units & 99.71 \\
        & Insulin - Novolog         & 4.95    & 3.15    & units & 99.74 \\
        & Insulin - Humalog 75/25   & 14.82   & 6.68    & units & 99.94 \\
        \hline
        \multirow{6}{*}{Opioid Analgesics} 
        & Fentanyl                  & 45.02   & 24.31   & mcg   & 62.03 \\
        & Fentanyl (Concentrate)    & 0.59    & 0.85    & mg    & 68.36 \\
        & Hydromorphone (Dilaudid)  & 0.82    & 2.02    & mg    & 83.97 \\
        & Morphine Sulfate          & 2.89    & 5.77    & mg    & 83.85 \\
        & Meperidine (Demerol)      & 13.84   & 3.88    & mg    & 98.75 \\
        & Methadone Hydrochloride   & 13.75   & 8.71    & mg    & 99.85 \\
        \hline
        \multirow{7}{*}{Sedatives and Anesthetics} 
        & Propofol                  & 294.35  & 341.89  & mg    & 54.55 \\
        & Midazolam (Versed)        & 8.60    & 18.76   & mg    & 78.26 \\
        & Dexmedetomidine (Precedex)& 90.21   & 101.66  & mcg   & 91.29 \\
        & Lorazepam (Ativan)        & 1.18    & 1.14    & mg    & 87.72 \\
        & Ketamine                  & 58.58   & 38.57   & mg    & 99.76 \\
        & Diazepam (Valium)         & 7.20    & 4.57    & mg    & 99.72 \\
        & Pentobarbital             & 508.17  & 749.81  & mg    & 99.94 \\
        \hline
        Thrombolytics & Alteplase (TPA) & 20.68   & 24.19   & mg    & 99.78 \\
        \hline
        \multirow{8}{*}{Vasopressors and Inotropes} 
        & Norepinephrine            & 1.21    & 1.98    & mg    & 74.43 \\
        & Phenylephrine             & 7.93    & 15.50   & mg    & 74.93 \\
        & Epinephrine               & 0.87    & 1.81    & mg    & 95.28 \\
        & Dopamine                  & 83.34   & 117.36  & mg    & 96.44 \\
        & Vasopressin               & 18.72   & 20.28   & units & 93.20 \\
        & Milrinone                 & 11.06   & 7.48    & mg    & 97.82 \\
        & Dobutamine                & 110.95  & 109.13  & mg    & 98.37 \\
        & Isuprel                   & 0.39    & 0.47    & mg    & 99.91 \\
        \bottomrule
    \end{tabularx}
\end{table}

\FloatBarrier
\subsection*{Results Phenotyping per Class}

\begin{table}[ht]
    \centering
    \scriptsize 
    \renewcommand{\arraystretch}{1.2}
    \caption{AUROC values for all methods and modality combinations for task \textit{phenotyping} per class. Additionally, the prevalence of each class in train- and testset is reported. All = $C|M|X|E$.}
    \label{tab:results-phenotyping}
    \begin{tabularx}{\linewidth}{
    p{0.27\linewidth}|
    C{0.045\linewidth}C{0.045\linewidth}|
    C{0.045\linewidth}C{0.045\linewidth}C{0.045\linewidth}|
    C{0.045\linewidth}C{0.045\linewidth}C{0.045\linewidth}|
    C{0.045\linewidth}C{0.045\linewidth}C{0.045\linewidth}C{0.045\linewidth}C{0.045\linewidth}C{0.045\linewidth}
    }
        \toprule
        \multirow{2}{*}{Phenotype} & \multicolumn{2}{c|}{Prevalence} & \multicolumn{3}{c|}{MeTra} & \multicolumn{3}{c|}{MedFuse} & \multicolumn{6}{c}{ViTiMM (ours)} \\ 
         & Train & Test & C & X & $C|X$ & C & X & $C|X$ & C & X & M & E & $C|X$ & All \\ 
        \midrule
        Acute and unspecified renal failure & 0.371 & 0.358 & 0.707 & 0.679 & 0.713 & 0.741 & 0.660 & 0.753 & 0.870 & 0.743 & 0.746 & 0.729 & 0.869 & 0.872 \\
        Acute cerebrovascular disease & 0.099 & 0.095 & 0.897 & 0.683 & 0.889 & 0.881 & 0.644 & 0.888 & 0.870 & 0.758 & 0.768 & 0.662 & 0.860 & 0.878 \\
        Acute myocardial infarction & 0.090 & 0.098 & 0.710 & 0.650 & 0.700 & 0.724 & 0.644 & 0.717 & 0.891 & 0.805 & 0.773 & 0.704 & 0.912 & 0.903 \\
        Cardiac dysrhythmias & 0.413 & 0.405 & 0.603 & 0.666 & 0.643 & 0.618 & 0.654 & 0.681 & 0.703 & 0.698 & 0.705 & 0.676 & 0.709 & 0.705 \\
        Chronic kidney disease & 0.250 & 0.243 & 0.711 & 0.707 & 0.739 & 0.737 & 0.687 & 0.776 & 0.903 & 0.757 & 0.745 & 0.739 & 0.906 & 0.907 \\
        COPD and bronchiectasis & 0.172 & 0.151 & 0.623 & 0.726 & 0.720 & 0.632 & 0.671 & 0.690 & 0.695 & 0.751 & 0.652 & 0.588 & 0.754 & 0.742 \\
        Complications of procedures & 0.264 & 0.277 & 0.670 & 0.606 & 0.653 & 0.660 & 0.546 & 0.664 & 0.666 & 0.648 & 0.662 & 0.646 & 0.666 & 0.677 \\
        Conduction disorders & 0.111 & 0.119 & 0.680 & 0.799 & 0.789 & 0.626 & 0.707 & 0.796 & 0.752 & 0.850 & 0.751 & 0.801 & 0.847 & 0.874 \\
        Congestive heart failure & 0.336 & 0.307 & 0.710 & 0.743 & 0.759 & 0.731 & 0.726 & 0.798 & 0.816 & 0.813 & 0.807 & 0.815 & 0.828 & 0.819 \\
        Coronary atherosclerosis & 0.317 & 0.326 & 0.727 & 0.734 & 0.745 & 0.725 & 0.681 & 0.743 & 0.807 & 0.788 & 0.779 & 0.726 & 0.831 & 0.809 \\
        Diabetes w/ complications & 0.119 & 0.117 & 0.810 & 0.583 & 0.817 & 0.829 & 0.567 & 0.827 & 0.826 & 0.809 & 0.819 & 0.818 & 0.823 & 0.852 \\
        Diabetes w/o complication & 0.207 & 0.221 & 0.673 & 0.598 & 0.692 & 0.699 & 0.547 & 0.717 & 0.728 & 0.706 & 0.720 & 0.681 & 0.714 & 0.719 \\
        Disorders of lipid metabolism & 0.397 & 0.414 & 0.637 & 0.669 & 0.654 & 0.663 & 0.630 & 0.675 & 0.688 & 0.666 & 0.682 & 0.656 & 0.687 & 0.684 \\
        Essential hypertension & 0.435 & 0.437 & 0.598 & 0.613 & 0.645 & 0.643 & 0.611 & 0.681 & 0.728 & 0.662 & 0.665 & 0.633 & 0.725 & 0.726 \\
        Fluid and electrolyte disorders & 0.512 & 0.513 & 0.690 & 0.638 & 0.683 & 0.682 & 0.632 & 0.681 & 0.729 & 0.664 & 0.654 & 0.660 & 0.728 & 0.718 \\
        Gastrointestinal hemorrhage & 0.072 & 0.071 & 0.667 & 0.594 & 0.663 & 0.697 & 0.597 & 0.690 & 0.741 & 0.664 & 0.660 & 0.579 & 0.727 & 0.777 \\
        Hypertension w/ complications & 0.225 & 0.221 & 0.686 & 0.686 & 0.723 & 0.719 & 0.691 & 0.770 & 0.886 & 0.752 & 0.749 & 0.725 & 0.882 & 0.882 \\
        Other liver diseases & 0.179 & 0.183 & 0.665 & 0.660 & 0.695 & 0.673 & 0.686 & 0.719 & 0.775 & 0.714 & 0.676 & 0.614 & 0.779 & 0.783 \\
        Other lower respiratory disease & 0.140 & 0.144 & 0.532 & 0.604 & 0.546 & 0.566 & 0.551 & 0.580 & 0.555 & 0.582 & 0.528 & 0.462 & 0.576 & 0.573 \\
        Other upper respiratory disease & 0.074 & 0.058 & 0.675 & 0.560 & 0.663 & 0.703 & 0.544 & 0.721 & 0.733 & 0.630 & 0.657 & 0.574 & 0.688 & 0.734 \\
        Pleurisy; pneumothorax; collapse & 0.113 & 0.101 & 0.574 & 0.668 & 0.604 & 0.605 & 0.661 & 0.656 & 0.541 & 0.649 & 0.568 & 0.596 & 0.638 & 0.658 \\
        Pneumonia (except TB/STI) & 0.228 & 0.232 & 0.696 & 0.677 & 0.727 & 0.721 & 0.633 & 0.739 & 0.722 & 0.707 & 0.661 & 0.656 & 0.732 & 0.742 \\
        Respiratory failure & 0.350 & 0.360 & 0.792 & 0.731 & 0.801 & 0.785 & 0.675 & 0.788 & 0.804 & 0.757 & 0.746 & 0.724 & 0.805 & 0.809 \\
        Septicemia & 0.265 & 0.264 & 0.760 & 0.685 & 0.764 & 0.762 & 0.675 & 0.766 & 0.817 & 0.789 & 0.773 & 0.781 & 0.821 & 0.821 \\
        Shock & 0.224 & 0.217 & 0.785 & 0.707 & 0.782 & 0.809 & 0.676 & 0.807 & 0.889 & 0.870 & 0.873 & 0.874 & 0.892 & 0.891 \\
        \bottomrule
    \end{tabularx}
\end{table}

\FloatBarrier
\subsection*{Significance Tests}
\FloatBarrier

\begin{table}[H]
    \caption{
    To assess the statistical significance for predicting in-hospital mortality, we conducted pairwise comparisons using the paired samples t-test~\cite{ross2017ttest}.}
    \label{tab:p-values-inhospital-mortality}
    \begin{tabularx}{\linewidth}{
        p{0.09\linewidth}p{0.04\linewidth}|
        C{0.07\linewidth}C{0.07\linewidth}|
        C{0.07\linewidth}C{0.07\linewidth}C{0.07\linewidth}|
        C{0.07\linewidth}C{0.07\linewidth}C{0.07\linewidth}C{0.07\linewidth}C{0.07\linewidth}C{0.07\linewidth}
    }
        \toprule
        & & \multicolumn{2}{c|}{MeTra} & \multicolumn{3}{c|}{MedFuse} & \multicolumn{6}{c}{ViTiMM} \\
        & & X & $C|X$ & C & X & $C|X$ & C & X & M & E & $C|X$ & $C|M|X|E$ \\
        \midrule
        \multirow{3}{*}{MeTra} & C & {\scriptsize $<0.001$} & {\scriptsize $<0.001$} & {\scriptsize $<0.001$} & 0.738 & 0.261 & {\scriptsize $<0.001$} & {\scriptsize $<0.001$} & {\scriptsize $<0.001$} & {\scriptsize $<0.001$} & {\scriptsize $<0.001$} & {\scriptsize $<0.001$} \\
        & X & - & {\scriptsize $<0.001$} & {\scriptsize $<0.001$} & {\scriptsize $<0.001$} & {\scriptsize $<0.001$} & {\scriptsize $<0.001$} & {\scriptsize $<0.001$} & {\scriptsize $<0.001$} & {\scriptsize $<0.001$} & {\scriptsize $<0.001$} & {\scriptsize $<0.001$} \\
        & $C|X$ &  & - & 0.444 & {\scriptsize $<0.001$} & {\scriptsize $<0.001$} & {\scriptsize $<0.001$} & {\scriptsize $<0.001$} & {\scriptsize $<0.001$} & {\scriptsize $<0.001$} & {\scriptsize $<0.001$} & {\scriptsize $<0.001$} \\
        \midrule
        \multirow{3}{*}{MedFuse} & C &  &  & - & {\scriptsize $<0.001$} & {\scriptsize $<0.001$} & {\scriptsize $<0.001$} & {\scriptsize $<0.001$} & {\scriptsize $<0.001$} & {\scriptsize $<0.001$} & {\scriptsize $<0.001$} & {\scriptsize $<0.001$} \\
        & X &  &  &  & - & 0.253 & {\scriptsize $<0.001$} & {\scriptsize $<0.001$} & {\scriptsize $<0.001$} & {\scriptsize $<0.001$} & {\scriptsize $<0.001$} & {\scriptsize $<0.001$} \\
        & $C|X$ &  &  &  &  & - & {\scriptsize $<0.001$} & {\scriptsize $<0.001$} & {\scriptsize $<0.001$} & {\scriptsize $<0.001$} & {\scriptsize $<0.001$} & {\scriptsize $<0.001$} \\
        \midrule
        \multirow{5}{*}{ViTiMM} & C &  &  &  &  &  & - & 0.104 & {\scriptsize $<0.001$} & {\scriptsize $<0.001$} & {\scriptsize $<0.001$} & {\scriptsize $<0.001$} \\
        & X &  &  &  &  &  &  & - & {\scriptsize $<0.001$} & {\scriptsize $<0.001$} & {\scriptsize $<0.001$} & {\scriptsize $<0.001$} \\
        & M &  &  &  &  &  &  &  & - & 0.143 & {\scriptsize $<0.001$} & {\scriptsize $<0.001$} \\
        & E &  &  &  &  &  &  &  &  & - & {\scriptsize $<0.001$} & {\scriptsize $<0.001$} \\
        & $C|X$ &  &  &  &  &  &  &  &  &  & - & {\scriptsize $<0.001$} \\
        \bottomrule
    \end{tabularx}
\end{table}

\begin{table}[H]
    \caption{
    To assess the statistical significance for phenotyping, we conducted pairwise comparisons using the paired samples t-test~\cite{ross2017ttest}.}
    \label{tab:p-values-phenotyping}
    \begin{tabularx}{\linewidth}{
        p{0.09\linewidth}p{0.04\linewidth}|
        C{0.07\linewidth}C{0.07\linewidth}|
        C{0.07\linewidth}C{0.07\linewidth}C{0.07\linewidth}|
        C{0.07\linewidth}C{0.07\linewidth}C{0.07\linewidth}C{0.07\linewidth}C{0.07\linewidth}C{0.07\linewidth}
    }
        \toprule
        & & \multicolumn{2}{c|}{MeTra} & \multicolumn{3}{c|}{MedFuse} & \multicolumn{6}{c}{ViTiMM} \\
        & & X & $C|X$ & C & X & $C|X$ & C & X & M & E & $C|X$ & $C|M|X|E$ \\
        \midrule
        \multirow{3}{*}{MeTra} & C & {\scriptsize $<0.001$} & {\scriptsize $<0.001$} & {\scriptsize $<0.001$} & {\scriptsize $<0.001$} & {\scriptsize $<0.001$} & 0.005 & {\scriptsize $<0.001$} & {\scriptsize $<0.001$} & 0.004 & {\scriptsize $<0.001$} & {\scriptsize $<0.001$} \\
        & X & - & {\scriptsize $<0.001$} & {\scriptsize $<0.001$} & 0.717 & 0.362 & {\scriptsize $<0.001$} & {\scriptsize $<0.001$} & {\scriptsize $<0.001$} & {\scriptsize $<0.001$} & {\scriptsize $<0.001$} & {\scriptsize $<0.001$} \\
        & $C|X$ &  & - & {\scriptsize $<0.001$} & {\scriptsize $<0.001$} & {\scriptsize $<0.001$} & {\scriptsize $<0.001$} & {\scriptsize $<0.001$} & {\scriptsize $<0.001$} & {\scriptsize $<0.001$} & {\scriptsize $<0.001$} & 0.279 \\
        \midrule
        \multirow{3}{*}{MedFuse} & C &  &  & - & {\scriptsize $<0.001$} & {\scriptsize $<0.001$} & {\scriptsize $<0.001$} & {\scriptsize $<0.001$} & {\scriptsize $<0.001$} & {\scriptsize $<0.001$} & {\scriptsize $<0.001$} & {\scriptsize $<0.001$} \\
        & X &  &  &  & - & 0.111 & {\scriptsize $<0.001$} & {\scriptsize $<0.001$} & {\scriptsize $<0.001$} & {\scriptsize $<0.001$} & {\scriptsize $<0.001$} & {\scriptsize $<0.001$} \\
        & $C|X$ &  &  &  &  & - & {\scriptsize $<0.001$} & {\scriptsize $<0.001$} & {\scriptsize $<0.001$} & {\scriptsize $<0.001$} & {\scriptsize $<0.001$} & {\scriptsize $<0.001$} \\
        \midrule
        \multirow{5}{*}{ViTiMM} & C &  &  &  &  &  & - & 0.003 & {\scriptsize $<0.001$} & {\scriptsize $<0.001$} & {\scriptsize $<0.001$} & {\scriptsize $<0.001$} \\
        & X &  &  &  &  &  &  & - & 0.283 & {\scriptsize $<0.001$} & {\scriptsize $<0.001$} & {\scriptsize $<0.001$} \\
        & M &  &  &  &  &  &  &  & - & {\scriptsize $<0.001$} & {\scriptsize $<0.001$} & {\scriptsize $<0.001$} \\
        & E &  &  &  &  &  &  &  &  & - & 0.001 & 0.062 \\
        & $C|X$ &  &  &  &  &  &  &  &  &  & - & {\scriptsize $<0.001$} \\
        \bottomrule
    \end{tabularx}
\end{table}

\newpage
\FloatBarrier
\subsection*{Attention Visualization}
\FloatBarrier

\begin{figure}[]
    \centering
    \includegraphics[width=0.65\linewidth, trim=0 2cm 0 0, clip]{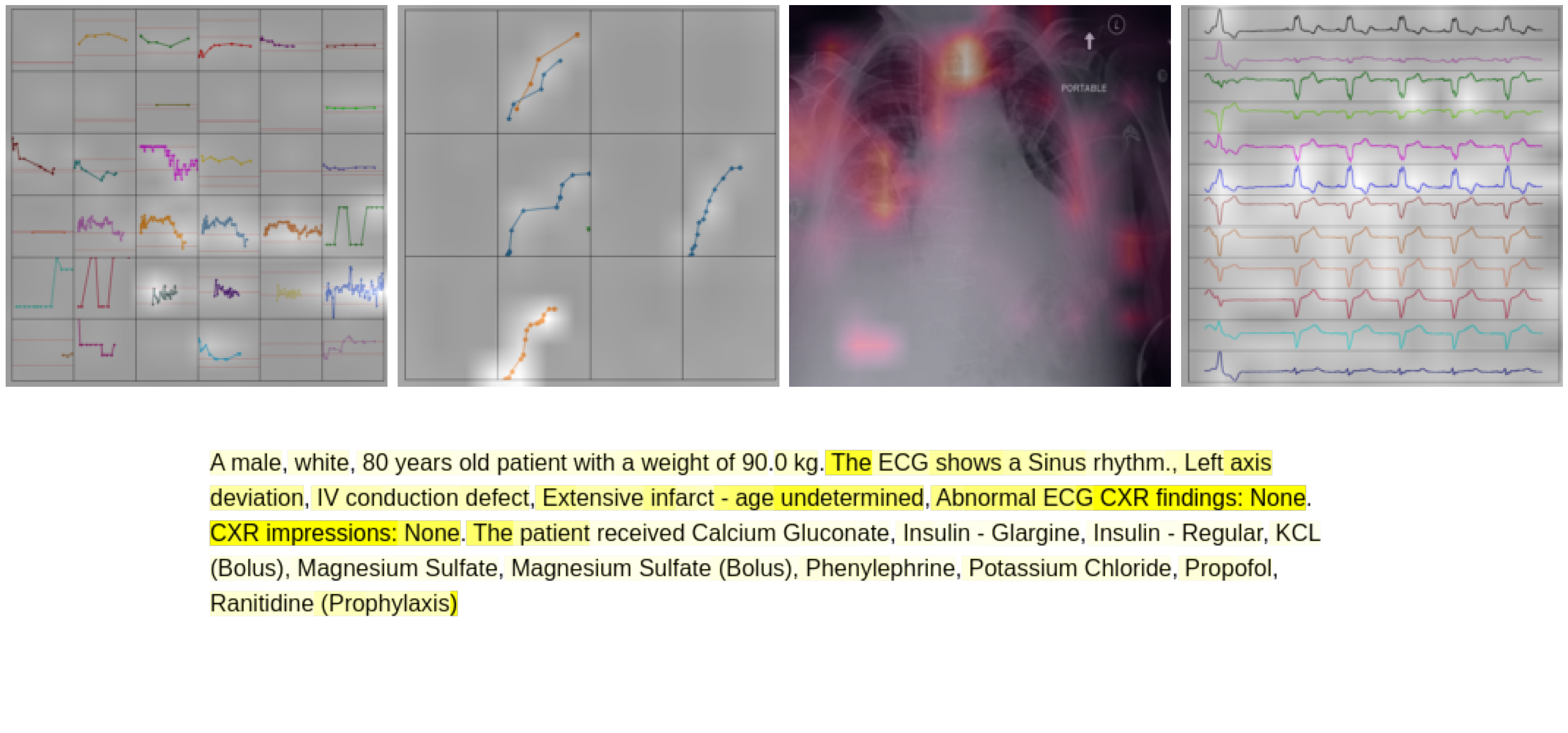}
    \includegraphics[width=0.65\linewidth, trim=0 2cm 0 0, clip]{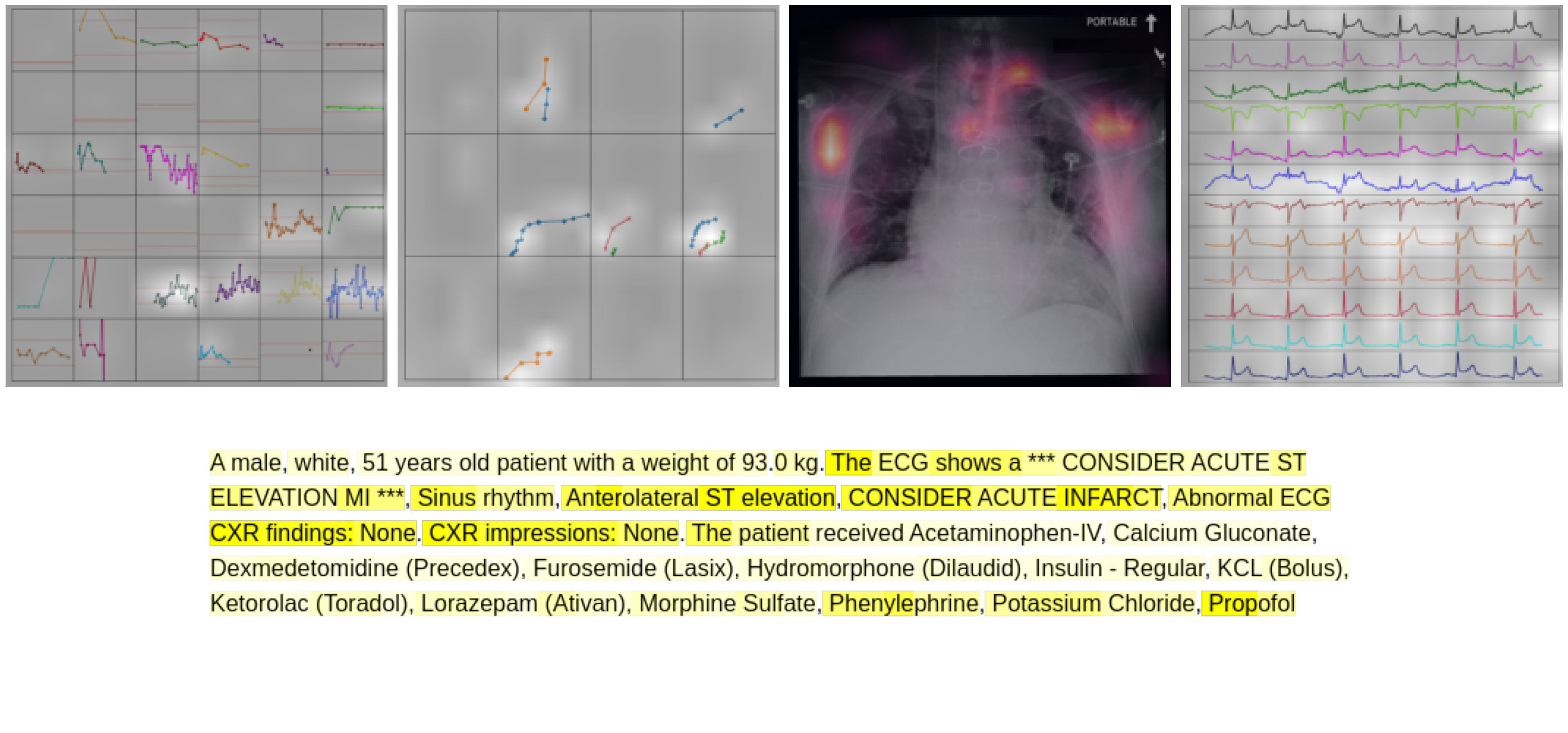}
    \includegraphics[width=0.65\linewidth, trim=0 5cm 0 0, clip]{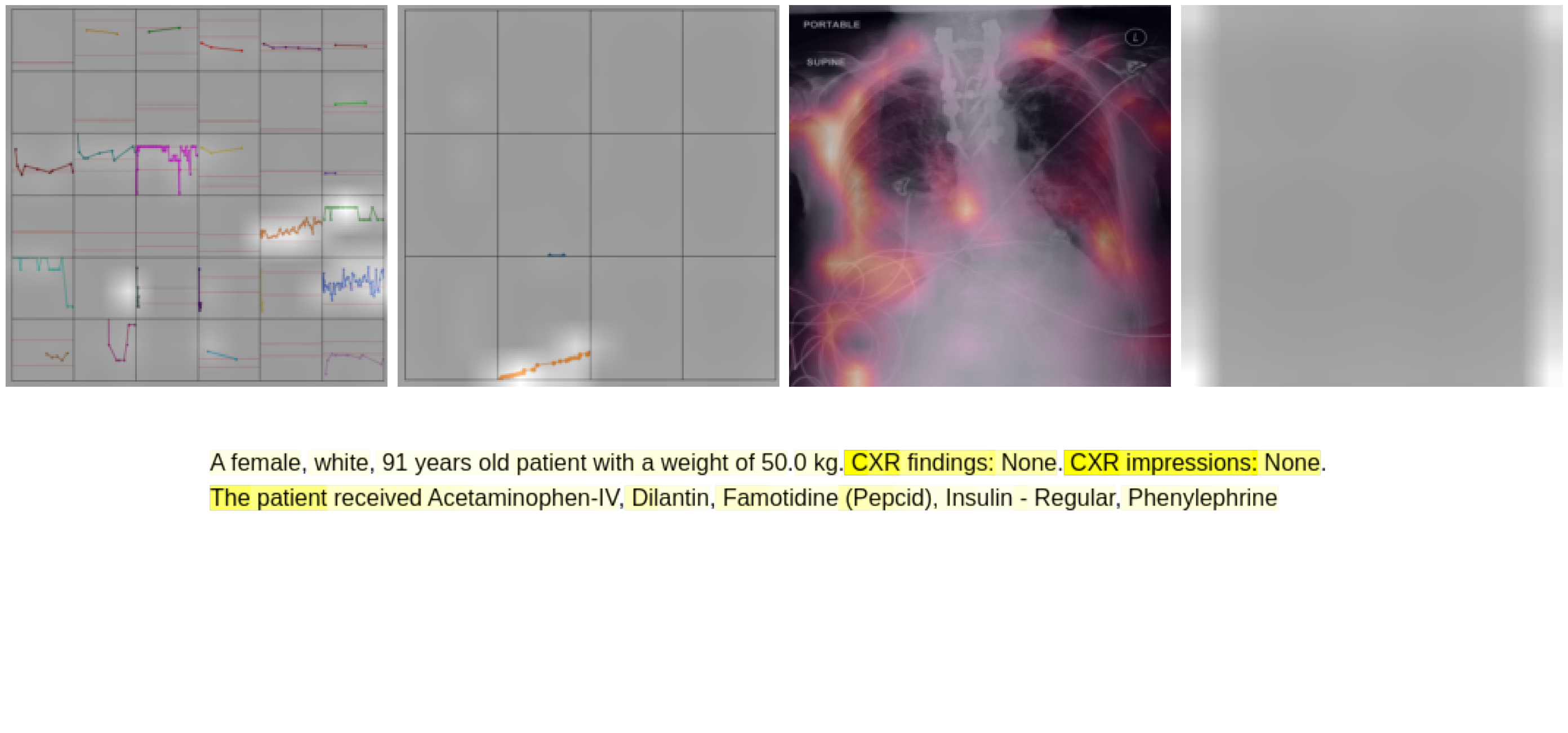}
    \caption{Attention Maps of ViT for \textit{In-hospital Mortality} prediction for \texttt{hadm\_ids\,30009123,\,30090650,\,30055897}.
    The patients survived the hospital stay, which the model correctly predicted in the top two cases but failed to do in the bottom one. 
    However, with a predicted probability of 0.602, the model reflects a degree of uncertainty.}
    \label{fig:attention-mortality-0}
\end{figure}

\begin{figure}[]
    \centering
    \includegraphics[width=0.7\linewidth, trim=0 2cm 0 0, clip]{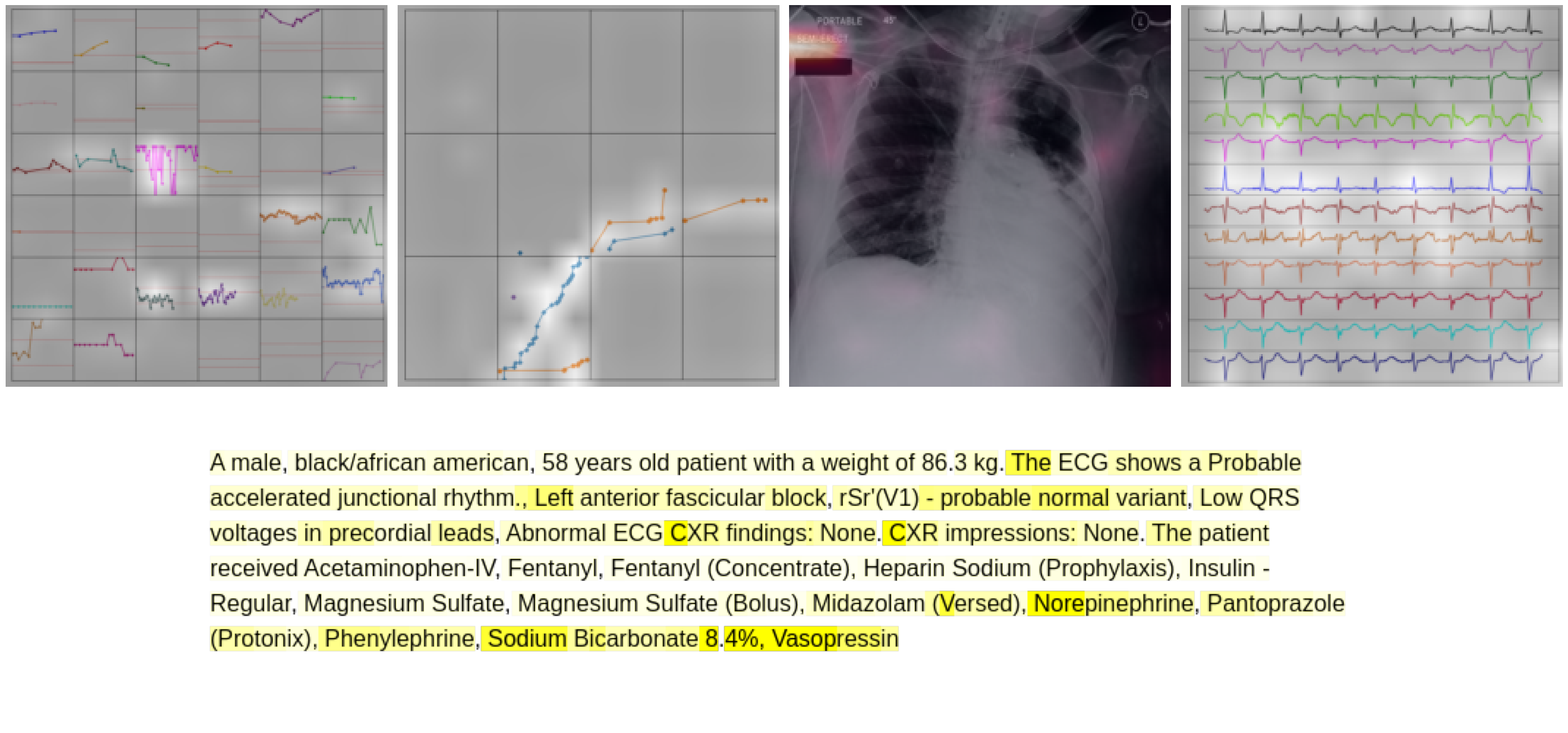} 
    \includegraphics[width=0.7\linewidth, trim=0 0 0 0, clip]{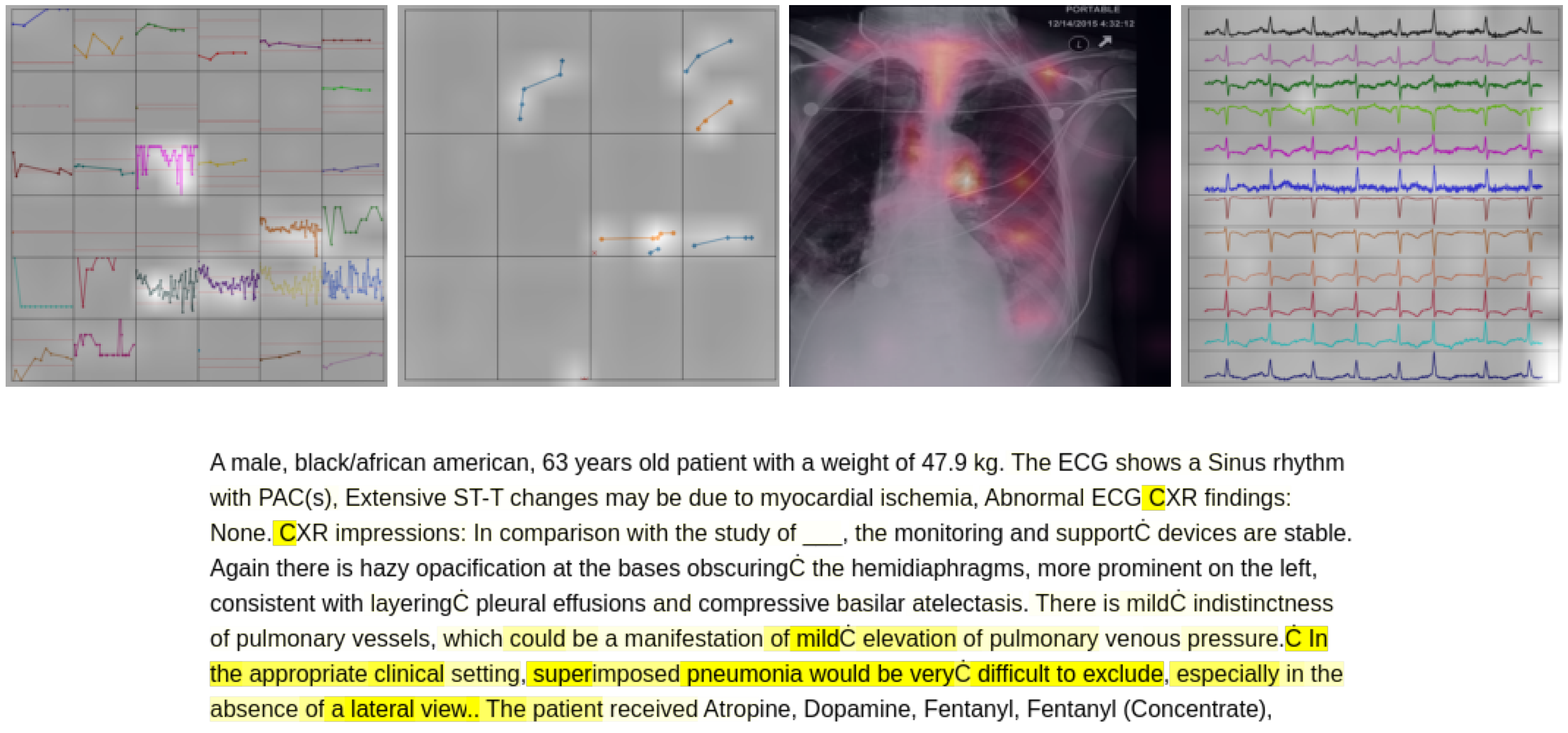}
    \includegraphics[width=0.7\linewidth, trim=0 2cm 0 0, clip]{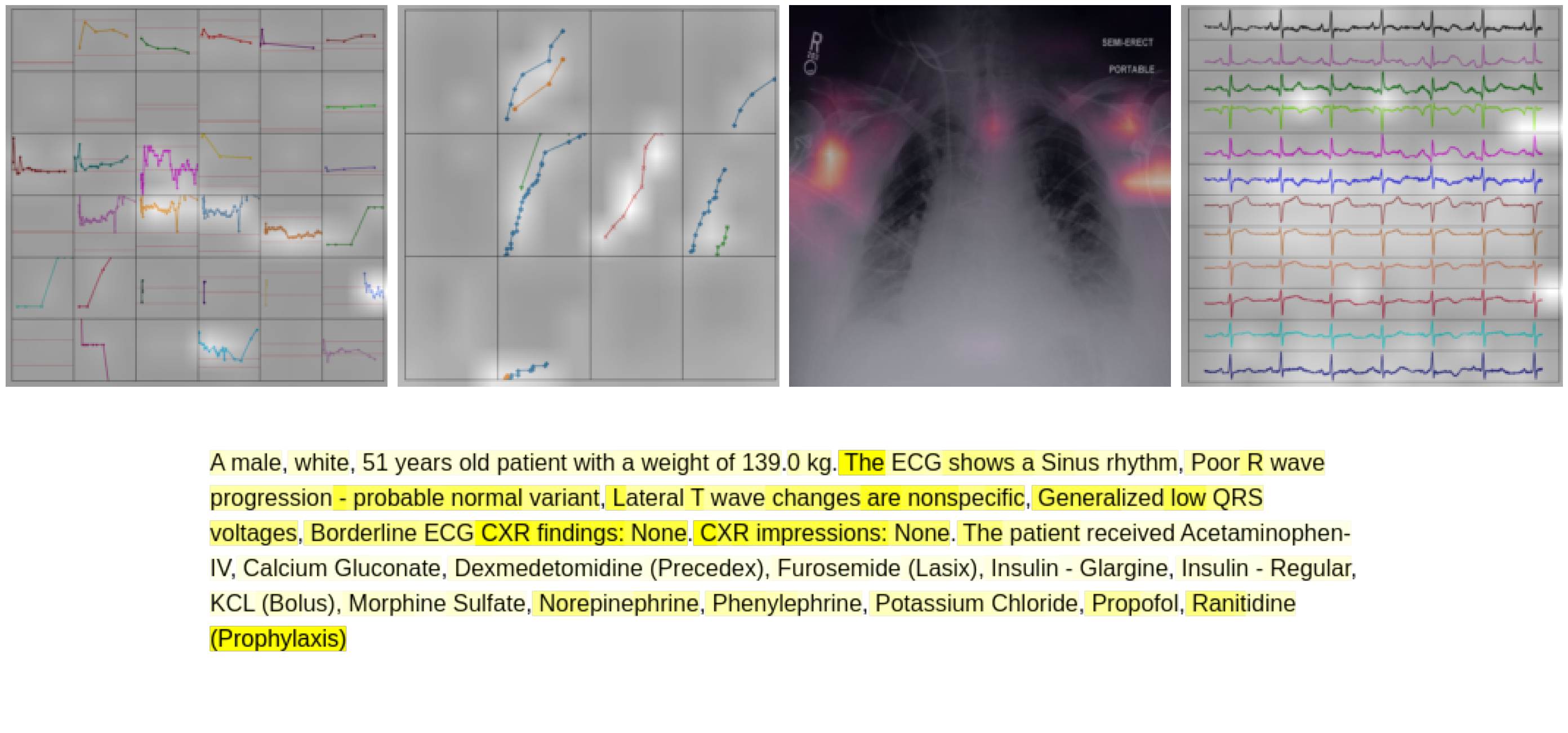}
    \caption{Attention Maps of ViT for \textit{In-hospital Mortality} prediction for \texttt{hadm\_ids\,30204754,\,30397772,\,30596506}.
    The patients survived the hospital stay, which the model correctly predicted in the top two cases but again failed to do in the bottom one.}
    \label{fig:attention-mortality-1}
\end{figure}

\end{document}